\documentclass{article}

\usepackage{fullpage}
\usepackage[numbers]{natbib}
\usepackage{authblk}

\usepackage[utf8]{inputenc} %
\usepackage[T1]{fontenc}    %
\usepackage{hyperref}       %
\usepackage{url}            %
\usepackage{booktabs}       %
\usepackage{amsfonts}       %
\usepackage{nicefrac}       %
\usepackage{microtype}      %
\usepackage{xcolor}         %

\usepackage{amsmath}
\usepackage{amsthm}
\usepackage{dsfont}
\usepackage{amssymb}
\usepackage{amscd}
\usepackage{mathtools}
\mathtoolsset{showonlyrefs=true}
\usepackage{enumerate}
\usepackage{authblk}
\usepackage{ bbold }
\usepackage[ruled,vlined]{algorithm2e}
\usepackage{wrapfig}

\newtheorem{theorem}{Theorem}
\newtheorem{prop}{Property}

\newcommand{\argmax}{\operatorname*{argmax}} 

\newcommand{\sm}{\operatorname*{softmax}}
\newcommand{\kl}[2]{\operatorname*{KL}(#1||#2)}

\newcommand{\pr}{\operatorname*{Pr}} 
\newcommand{\R}{\mathbb{R}}
\newcommand{\hc}{\mathcal{H}}
\newcommand{\rc}{\mathcal{R}}
\newcommand{\states}{\mathcal{S}}
\newcommand{\actions}{\mathcal{A}}

\newcommand{\E}{\mathbb{E}}

\newcommand\myeq[2]{\stackrel{\mathclap{(#1)}}{#2}}

\title{Concave Utility Reinforcement Learning:\\ the Mean-Field Game Viewpoint}

\author[1]{Matthieu Geist}
\author[2]{Julien Pérolat}
\author[1]{Mathieu Laurière}
\author[2]{Romuald Elie}
\author[3]{Sarah Perrin}
\author[1]{Olivier Bachem}
\author[2]{Rémi Munos}
\author[1]{Olivier Pietquin}

\affil[1]{Google Research, Brain Team}
\affil[2]{DeepMind Paris}
\affil[3]{Univ. Lille, CNRS, Inria, Centrale Lille, UMR 9189 CRIStAL}

\date{}

\begin{document}

\maketitle

\begin{abstract}
  Concave Utility Reinforcement Learning (CURL) extends RL from linear to concave utilities in the occupancy measure induced by the agent's policy. This encompasses not only RL but also imitation learning and exploration, among others. Yet, this more general paradigm invalidates the classical Bellman equations, and calls for new algorithms. Mean-field Games (MFGs) are a continuous approximation of many-agent RL. They consider the limit case of a continuous distribution of identical agents, anonymous with symmetric interests, and reduce the problem to the study of a single representative agent in interaction with the full population. Our core contribution consists in showing that CURL is a subclass of MFGs. We think this important to bridge together both communities. It also allows to shed light on aspects of both fields: we show the equivalence between concavity in CURL and  monotonicity in the associated MFG, between optimality conditions in CURL and Nash equilibrium in MFG, or that Fictitious Play (FP) for this class of MFGs is simply Frank-Wolfe, bringing the first convergence rate for discrete-time FP for MFGs. We also experimentally demonstrate that, using algorithms recently introduced for solving MFGs, we can address the CURL problem more efficiently.
\end{abstract}

\section{Introduction}
\label{sec:introduction}

\looseness=-1 Reinforcement Learning (RL) aims at finding a policy maximizing the expected discounted cumulative reward along trajectories generated by the policy. This objective can be rewritten as an inner product between the occupancy measure induced by the policy (\textit{i.e.}, the discounted distribution of state-action pairs it visits) and a policy-independent reward for each state-action pair. Yet, several recently investigated problems aim at optimizing a more general function of this occupancy measure:  \citet{hazan2019provably} propose pure exploration in RL as maximizing the entropy of the policy-induced state occupancy measure, and \citet{ghasemipour2020divergence} showed that many imitation learning (IL) algorithms minimize an $f$-divergence between the policy-induced state-action occupancy measures of the agent and the expert. Recently, \citet{hazan2019provably} or \citet{zhang2020variational} abstracted such problems as Concave (or general) Utility Reinforcement Learning (CURL), that aims at maximizing a concave function of the occupancy measure, over policies generating them.  %
This calls for new algorithms since usual RL approaches do not readily apply.

\looseness=-1
Mean-field games (MFGs) have been introduced concurrently by~\citet{lasry2007mean} and~\citet{huang2006large}. They address sequential decision problems involving a large population of anonymous agents, with symmetric interests. They do so by considering the limiting case of a continuous distribution of agents, which allows reducing the original problem to the study of a single representative agent through its interaction with the full population. MFGs have been originally studied in Mathematics, with solutions typically involving coupled differential equations, but has recently seen a surge of interest in Machine Learning, \textit{e.g.}~\citep{yang2018mean,guo2019learning,perrin2020fictitious,cao2020connecting}. %

Our core contribution consists in showing that CURL is indeed a subclass of MFGs. We think this to be of interest to bridge together both communities and shed some light on connections between both fields. For example, we show an equivalence between the concavity of the CURL function and the  monotonicity of the corresponding MFG (monotonicity being a sufficient condition to guarantee the existence of a Nash equilibrium). We also show an equivalence between the optimality conditions of the CURL problem (hence its maximizer) and the exploitability of the corresponding MFG (characterizing the Nash equilibrium). 
We also discuss algorithms for both fields. We show that the seminal approach of~\citet{hazan2019provably} for CURL is nothing else than the classic Fictitious Play (FP) for MFGs~\citep{cardaliaguet2017learning,perrin2020fictitious}. We also show that both are in fact instances of Frank-Wolfe~\citep{frank1956algorithm} applied to the CURL problem. This readily provides the first (to the best of our knowledge) convergence rate for discrete-time FP applied to an MFG, consistent with the one hypothesized by~\citet{perrin2020fictitious} while studying continuous-time FP. This connection between fields also allows applying MFGs algorithms to CURL. In a set of numerical illustrations, we apply the recently introduced Online Mirror Descent (OMD) for MFGs~\citep{perolat2021scaling} to CURL, which appears to converge much faster than classic FP. %

\section{Background and Related Works}
\label{sec:background}

We write $\Delta_X$ the set of probability distributions over a finite set $X$ and $Y^X$ the set of applications from $X$ to the set $Y$. For $f_1,f_2\in\R^X$, we write the dot product $\langle f_1, f_2\rangle = \sum_{x\in X} f_1(x) f_2(x)$.
RL is usually formalized using Markov Decision Processes (MDPs). An MDP is a tuple $\{\states,\actions, P, r, \gamma,\rho_0\}$ with $\states$ the finite state space, $\actions$ the finite action space, $P\in\Delta^{\states\times\actions}_\states$ the Markovian transition kernel, $r\in\R^{\states\times\actions}$ the (bounded) reward function, $\gamma\in(0,1)$ the discount factor and $\rho_0\in\Delta_{\states}$ the initial state distribution.
A policy $\pi\in\Delta_{\actions}^\states$ associates to each state a distribution over actions. Its (state-action) value function is $q_{\pi,r}(s,a) = \E_\pi[\sum_{t\geq 0}\gamma^t r(S_t,A_t)|S_0=s, A_0=a]$, with $\E_\pi$ the expectation over trajectories induced by the policy $\pi$ (and the dynamics $P$). RL aims at finding a policy maximizing the value for any state-action pair, or as considered here for initial states sampled according to $\rho_0$. Writing $\mu_{0,\pi} = \rho_0\pi\in\Delta_{\states\times\actions}$ ($\mu_{0,\pi}(s,a) = \rho_0(s) \pi(a|s)$), the RL problem can thus be framed as (the factor $(1-\gamma)$ accounting for later normalization) $\max_{\pi\in\Delta_\actions^\states} J_\text{rl}(\pi)$ with
\begin{equation}
    J_\text{rl}(\pi) = (1-\gamma) \E_{(S,A)\sim\mu_{0,\pi}}[q_{\pi,r}(S,A)] = (1-\gamma)\langle \mu_{0,\pi}, q_{\pi,r}\rangle.
\end{equation}
Define the policy kernel $P_\pi$ as $P_\pi(s',a'|s,a) = P(s'|s,a)\pi(a'|s')$. The value function is the unique fixed point of the Bellman equation~\citep{puterman2014markov}, $q_{\pi,r} = r + \gamma P_\pi q_{\pi,r}$, or equivalently $q_{\pi,r} = (I-\gamma P_\pi)^{-1}r$, with $I$ the identity. We'll use the discounted occupancy measure induced by $\pi$, $\mu_\pi\in\Delta_{\states\times\actions}$, defined as $\mu_\pi(s,a) = (1-\gamma)\sum_{t\geq 0} \gamma^t \pr(S_t=s,A_t=a|S_0\sim\rho_0, \pi)$, or for short $\mu_\pi = (1-\gamma)(I-\gamma P_\pi^\top)^{-1} \mu_{0,\pi}$. %
With this additional notation, the RL objective function is %
\begin{equation}
    J_\text{rl}(\pi) = \langle \mu_{\pi}, r\rangle.
    \label{eq:rl_problem}
\end{equation}
We clearly see the linearity of the RL objective in terms of the policy-induced occupancy measure (but not in terms of the policy). As it will be useful later, we also introduce the set $M\subset \Delta_{\states\times\actions}$ of state-action distributions satisfying the Bellman flow (\textit{e.g.},~\citep{zhang2020variational}):
\begin{equation}
    M = \left\{\mu\in\Delta_{\states\times\actions}| \sum_{a\in\actions} \mu(\cdot,a) = (1-\gamma)\rho_0 + \gamma P^\top \mu\right\}. 
    \label{eq:def_M}
\end{equation}
For any  $\mu\in\Delta_{\states\times\actions}$, we write $\rho\in\Delta_\states$ its marginal over states:
\begin{equation}
    \forall s\in\states, \; \rho(s) = \sum_{a\in\actions} \mu(s,a).
    \label{eq:def_rho}
\end{equation}
For any policy $\pi\in\Delta_\actions^\states$, we have $\mu_\pi\in M$, and for any $\mu\in M$, defining $\pi(a|s) = \frac{\mu(s,a)}{\rho(s)}$, %
we have $\mu_\pi = \mu$. Thus, the RL problem defined Eq.~\eqref{eq:rl_problem} 
is %
a linear program, $\max_{\pi\in\Delta_\actions^\states} \langle \mu_\pi, r\rangle = \max_{\mu\in M} \langle \mu, r\rangle$. 
 
\subsection{Concave Utility Reinforcement Learning}
\label{subsec:curl}

Let $F\in\R^{\Delta_{\states\times\actions}}$ be a concave function, CURL~\citep{hazan2019provably} aims at solving
\begin{equation}
    \max_{\pi\in\Delta_\actions^\states} F(\mu_\pi).
    \label{eq:curl_problem}
\end{equation}
As discussed above, RL is a special case with $F(\mu) = \langle  \mu, r\rangle$, but this representation also subsumes other relevant related problems. We present below a probably non-exhaustive list of them.

\textbf{Exploration.} \citet{hazan2019provably} formulated the pure exploration problem as finding the policy maximizing the entropy of the induced state occupancy measure, corresponding to $F(\mu) = - \langle \rho, \ln \rho\rangle$ (recall the definition of $\rho$ from Eq.~\eqref{eq:def_rho}). One could consider the entropy of the state-action occupancy measure, $F(\mu) = -\langle \mu, \ln \mu\rangle$, but it would probably make less sense from an exploration perspective (we usually want to visit all states evenly, without caring about actions themselves). This could also be combined with the maximization of a cumulative reward, $F(\mu) = \langle \mu,r\rangle - \lambda \langle \rho, \ln\rho\rangle$, with $\lambda$ balancing task resolution and exploration.

\textbf{Divergence minimization.} For a given $f$-divergence $D_f$, this task corresponds to $F(\mu) = -D_f(\mu||\mu_*)$ or $F(\mu) = -D_f(\rho||\rho_*)$, with $\mu_*$ (resp. $\rho_*$) a target occupancy measure. The first case encompasses many \textbf{imitation learning} approaches~\citep{ghasemipour2020divergence}. There, the target measure $\mu_*$ is given by trajectories of an expert policy to be imitated. The second case encompasses for example the recently introduced framework of \textbf{state-marginal matching}~\citep{lee2019efficient}, that itself generalizes entropy maximization. The core difference with imitation learning is whether the occupancy measure of interest is directly accessible or only through samples. As noted by~\citet{zhang2020variational}, this approach can be generalized to any distance (\textit{e.g.}, Wasserstein), as long as it is convex in $\mu$.

\looseness=-1
\textbf{Constrained MDPs.} Let $c\in\R^{\states\times\actions}$ be a cost function. A constrained MDP aims at maximizing the cumulative reward while keeping the cumulative cost below a threshold $\mathcal{C}$: $\max_{\pi} \langle \mu_\pi, r\rangle$ such that $\langle \mu_\pi, c\rangle \leq \mathcal{C}$. As observed by~\citet{zhang2020variational}, a relaxed formulation can be framed in the CURL framework, with $F(\mu) = \langle \mu, r\rangle - \lambda p(\langle \mu, c\rangle - \mathcal{C})$, where $p$ is a convex penalty function.

\textbf{Multi-objectif RL} deals with the joint optimization of multiple rewards. Different approaches can be envisioned, some of them being a special case of CURL. Let $r^1\dots r^K$ be $K$ reward functions, and write $v^k_\pi = \langle \mu_\pi, r^k\rangle$. %
Consider $g:\R^K\rightarrow\R$ a concave function. Some multi-objective RL approaches~\citep{agarwal2019reinforcement,cheung2019regret} aim at solving $\max_\pi g(v^1_\pi,\dots,v^K_\pi)$. As a composition of a linear and a concave functions, it is concave in $\mu_\pi$, and thus a CURL problem.

\textbf{Offline RL}~\citep{levine2020offline} aims at learning a policy from a fixed dataset, without interacting with the system. A classic approach consists in penalizing the learnt policy when deviating too much from the one that generated the data, through some implicit or explicit divergence between policies~\citep{levine2020offline}. It would also make sense to penalize the learnt policy to reach states deviating from the ones in the dataset. If this approach was, up to our knowledge, never considered in the literature, it could also be framed as a CURL problem, with for example $F(\mu) = \langle \mu, r\rangle - \lambda D_f(\rho||\rho_\text{data})$. A similar and practical approach has been considered in the online RL setting~\citep{touati2020stable}.

\subsection{Mean-Field Games}
\label{subsec:mfg}

As explained in Sec.~\ref{sec:introduction}, 
MFGs are a continuous approximation of large population multi-agent problems. Thus, a core question is generally to know how good this approximation is, depending on the population size. Interestingly, we only need distributions of agents for drawing the connections to CURL, so we will not address this question further.
Moreover, it means that, whenever we refer to a population or a distribution of agents, it indeed corresponds to the occupancy measure induced by a single policy (there is a single agent, \textit{per se}).
Generally speaking, both the dynamics and the reward can depend on the population, but for drawing connections to CURL we need only population-dependent rewards. The (restricted) setting we present next is inspired  by  \textit{e.g.}~\citep{perrin2020fictitious,perolat2021scaling}, other settings exist (some being discussed in more details in Sec.~\ref{subsec:otherAlgs}).

An MFG is here  a tuple $\{\states,\actions,P,\rc,\gamma,\rho_0\}$. %
Everything is as in MDPs, except for the reward function $\rc\in\R^{\states\times\actions\times\Delta_\states}$ that now also depends on the state-distribution of the population. For such a fixed distribution $\rho\in\R^\states$, we define the following MDP-like criterion:
\begin{align}
    J(\pi,\rho) &= (1-\gamma)\E_\pi[\sum_{t\geq 0} \gamma^t \rc(S_t,A_t,\rho)|S_0\sim\rho_0] 
    = \langle \mu_\pi, \rc(\cdot, \rho)\rangle.
    \label{eq:mfg_criterion}
\end{align}

\textbf{Exploitability.} The exploitability $\phi(\pi)$ of a policy $\pi$ quantifies the maximum gain a representative player can get by deviating its policy from the rest of the population still playing $\pi$,
\begin{equation}
    \phi(\pi) = \max_{\pi'\in\Delta_\actions^\states} J(\pi',\rho_\pi) - J(\pi,\rho_\pi).
\end{equation}
The maximizer of $J(\pi,\rho)$ for a fixed $\rho$ is called a \textbf{best response} (optimal policy for the associated MDP).

\textbf{Nash equilibrium.} A Nash equilibrium is a policy satisfying $\phi(\pi)=0$: there is nothing to gain from deviating from the population policy. The core problem of MFGs is to compute this Nash.%

\textbf{Monotonicity.} Following~\citet{lasry2007mean}, a game is said to be \textbf{monotone} if for any $\mu,\mu'\in\Delta_{\states\times\actions}$, we have (with $\rho$ and $\rho'$ being defined through Eq.~\eqref{eq:def_rho}, again)
$%
    \langle \mu - \mu', \rc(\cdot,\rho) - \rc(\cdot,\rho')\rangle \leq 0%
$. %
This ensures existence of a Nash. If moreover the inequality is strict whenever $\mu\neq\mu'$, the game is said to be \textbf{strictly  monotone}, which ensures uniqueness of the Nash~\citep{perolat2021scaling}.

\textbf{Separable reward.} This special case will be of interest later. The reward is said to be separable if it decomposes as $\rc(s,a,\rho) = \bar{r}(s,a) + \tilde{r}(s,\rho)$. If in addition the following monotonicity condition holds, for all $\rho\neq\rho' \in \Delta_\states$, $\langle \rho - \rho', \tilde{r}(\cdot,\rho) - \tilde{r}(\cdot,\rho')\rangle \leq 0$ (resp. $<0$), then the game is monotone (resp. strictly  monotone)~\citep{perolat2021scaling}.

We will also consider less usual and slightly more general MFGs (sometimes referred to as Extended MFGs in the literature~\cite{gomes2016extended}), where the reward now depends on the state-action distribution, rather than on the state-distribution, $\rc\in\R^{\states\times\actions\times\Delta_{\states\times\actions}}$. The MDP-like criterion is thus $J(\pi,\mu) = \langle \mu_\pi, \rc(\cdot,\mu)\rangle$. Similarly to the previous case, we will say that the game is  monotone (resp. strictly) if for any $\mu\neq\mu'\in\Delta_{\states\times\actions}$, we have $\langle \mu -\mu', \rc(\cdot,\mu) - \rc(\cdot,\mu')\rangle \leq 0$ (resp. $<0$). Existence or uniqueness of the Nash through this monotonicity is not readily available from the literature, as far as we know, at least in an MDP setting. Yet, it will be a direct consequence of~Sec.~\ref{sec:curl_as_mfg}.

\subsection{Related Works}
\label{subsec:related_works}

As far as we know, this work is the first to connect CURL with MFGs. We briefly review relevant works of each of these fields.

\looseness=-1
\textbf{CURL.} All problems subsumed by CURL benefit from a large body of literature. For example, many articles address the problem of exploration in RL, not necessarily through the lens of entropy maximization (even though links can be drawn between predicting-error-based exploration and entropy maximization~\citep{lee2019efficient}). As stated before, many IL algorithms minimize a divergence between occupancy measures. Yet, they do not do it directly, and rather optimize a variational bound (to get a saddle-point problem involving a difference of expectation, to avoid estimating the target density). The approach of~\citet{lee2019efficient} for marginal matching minimizes a KL divergence differently, but the resulting algorithm is equivalent to the one of~\citet{hazan2019provably}, which tackles more general problems. %
There are two approaches that
address the general CURL problem: the one of~\citet{hazan2019provably} and the one of~\citet{zhang2020variational}, discussed more extensively in Sec.~\ref{sec:algorithms}.
Concurrently to our work, \citet{zahavy2021reward} also tackle the general CURL problem, by transforming it into a saddle-point problem, thanks to the Legendre-Fenchel transform of the concave objective. They then adapt the meta-algorithm of~\citet{Abernethy2017} and show that many algorithms handling special cases of CURL can be derived from it, yet relying on an explicit form of the convex conjugate of $F$. Our approach is different and complementary: we frame CURL as an MFG.

\looseness=-1
\textbf{MFGs.} There is also a large body of literature on MFGs. We will focus on the part more closely related to machine learning. A seminal paper combining MFG with learning was proposed by~\citet{yang2018mean} who analyze the convergence of mean field Q-learning and Actor-Critic algorithms to Nash.  \citet{guo2019learning} also combine Q-learning with MFGs, through a fixed-point approach. %
\citet{elie2020convergence} studies the propagation of errors due to learning and its effect on the Nash equilibrium. \citet{perrin2020fictitious} studies FP, both theoretically and empirically, when combined with learning. \citet{perolat2021scaling} introduces a mirror descent approach to MFG. These are only a few examples among others~\citep{mguni2018decentralised,yang2018deep,subramanian2019reinforcement}. Connections between MFGs and GANs \cite{cao2020connecting} or Deep Learning architectures~\citep{di2021deep} have also been established.
We think that an interesting aspect of our contribution is that many progress made on the MFGs side could be readily beneficial to CURL, thanks to what will be presented in Sec.~\ref{subsec:otherAlgs} (but up to a compatible setting, this will be discussed in Sec.~\ref{sec:algorithms}).
The converse may also be true: advances in CURL can benefit (possibly a subclass of) MFGs. 
Another relevant part of the MFG literature are potential MFGs (\textit{e.g.},~\citep{lasry2007mean,cardaliaguet2017learning}), where the reward function derives from a potential function. Results similar to those that we state in Sec.~\ref{sec:curl_as_mfg} appear there (link between monotonicity and concavity, maximizer and Nash), but within a different paradigm (notably, not within the MDP setting, possibly deterministic dynamics, specified through differential equations, finite horizon or even static case, etc.), and with no link to (CU)RL (\textit{e.g.},~\citep[Prop.~1.10, Lemma~5.72]{carmona2018probabilistic}).

\section{CURL is an MFG}
\label{sec:curl_as_mfg}

Now, we frame the CURL problem~\eqref{eq:curl_problem} as an MFG. We handle first the general case $F(\mu)$, and then the special case $F(\mu) = \langle \mu, r\rangle + G(\rho)$, with possibly $r=0$, as it has a specific structure representative of many problems depicted in Sec.~\ref{subsec:curl}. %

\textbf{General Case.}
Here, we consider the general CURL problem~\eqref{eq:curl_problem}. We define a potential MFG with the same dynamics, discount and initial distribution, but with reward deriving from the potential $F$:
\begin{equation}
    \forall \mu\in\Delta_{\states\times\actions}, \; \rc(\cdot,\mu) = \nabla F(\mu)\in\R^{\states\times \actions}. \label{eq:reward_general}
\end{equation}
The next results link the concavity of $F$ to the monotonicity of the game, and maximizers of CURL to Nash equilibria of the  MFG.
\begin{prop}
\label{prop:general}
    The MFG defined Eq.~\eqref{eq:reward_general} is (strictly)  monotone if and only if $F$ is (strictly) concave. %
\end{prop}
\begin{proof}
    From basic convex optimization principles, we have
    \begin{align}
        \text{\small $F$ concave} 
        &\myeq{a}{\Leftrightarrow} 
        \forall \mu_1,\mu_2\in\Delta_{\states\times\actions}, \langle \mu_2 - \mu_1, \nabla F(\mu_2) - \nabla F(\mu_1)\rangle \leq 0 %
        \\
        &\myeq{b}{\Leftrightarrow}
        \forall \mu_1,\mu_2\in\Delta_{\states\times\actions}, \langle \mu_2 - \mu_1, \rc(\cdot,\mu_2) - \rc(\cdot,\mu_1)\rangle \leq 0
        \\
        &\myeq{c}{\Leftrightarrow} \text{ the MFG is  monotone,}
    \end{align}
    with \emph{(a)} by~\cite[Thm.~2.1.3]{nesterov1998introductory}, \emph{(b)} by Eq.~\eqref{eq:reward_general} and \emph{(c)} by definition.
    The equivalence between strict concavity and strict  monotonicity is  obtained by replacing inequalities by strict ones for $\mu_1\neq\mu_2$.
\end{proof}
\begin{theorem}
\label{thm:general}
    Assume that $F$ is concave. Any maximizer of the CURL problem~\eqref{eq:curl_problem} is a Nash equilibrium of the MFG defined through reward~\eqref{eq:reward_general}, and conversely any Nash is a maximizer.
\end{theorem}
\begin{proof}
    We have that $\max_{\pi\in\Delta_\actions^\states} F(\mu_\pi) = \max_{\mu\in M} F(\mu)$, with $M$ defined in Eq.~\eqref{eq:def_M}. The later is a concave program with linear constraints, so concavity of $F$ ensures the existence of a maximizer, satisfying the optimality condition. Also, as seen in Sec.~\ref{sec:background}, for any policy $\pi$, $\mu_\pi\in M$, and for any $\mu\in M$, there is an associated policy (the conditional on actions, $\mu(s,a)/\rho(s)$). Therefore, we have
    \begin{align}
        \text{$\pi$ maximizer} &\myeq{a}{\Leftrightarrow}
        \langle \nabla F(\mu_\pi), \mu' - \mu_\pi\rangle\leq 0, \;\forall \mu'\in M
        \\
        &\myeq{b}{\Leftrightarrow}
        \langle \rc(\cdot,\mu_\pi), \mu_{\pi'} - \mu_\pi\rangle \leq 0, \;\forall \pi'\in\Delta_{\actions}^\states 
        \\
        &\Leftrightarrow
        \langle \mu_\pi', \rc(\cdot,\mu_\pi)\rangle \leq \langle \mu_\pi, \rc(\cdot,\mu_\pi)\rangle, \;\forall \pi'\in\Delta_{\actions}^\states 
        \\
        &\myeq{c}{\Leftrightarrow}
        J(\pi',\mu_\pi) \leq J(\pi,\mu_\pi), \;\forall \pi'\in\Delta_{\actions}^\states %
        \\
        &\myeq{d}{\Leftrightarrow}
        \phi(\pi) = 0 %
        \Leftrightarrow \text{$\pi$ is Nash}, 
    \end{align}
    with \emph{(a)} by optimality conditions, \emph{(b)} by using the associated policy and Eq.~\eqref{eq:reward_general}, \emph{(c)} by def. of $J$ and \emph{(d}) by def. of exploitability.
\end{proof}
Interestingly, the proof of Thm.~\ref{thm:general} shows an equivalence between the optimality conditions for CURL (that provides a global optimum, even though $F(\mu_\pi)$ is not concave in $\pi$, thanks to the relationship between $\Delta_\actions^\states$ and $M$) and a null exploitability for MFG.
As stated Sec.~\ref{subsec:mfg}, existence or uniqueness of the Nash for this class of MFGs from (strict)  monotonicity is not readily available in the literature, in this MDP setting. Yet, by Prop.~\ref{prop:general}  monotonicity and concavity are equivalent, by Thm.~\ref{thm:general}  the set of CURL maximizers and that of MFG Nash equilibria are equal, so the result holds readily (uniqueness in $M$ thanks to strict concavity implying uniqueness in $\Delta_{\actions}^\states$ here).

\textbf{A Relevant Special Case.}
We consider CURL~\eqref{eq:curl_problem} when $F$ takes the form ($\rho$ as in Eq.~\eqref{eq:def_rho})
\begin{equation}
    F(\mu) = \langle \mu, r\rangle + G(\rho),
    \label{eq:def_F_special}
\end{equation}
with $r$ being possibly null and $G$ being a differentiable potential function. This is representative of a number of examples of Sec.~\ref{subsec:curl}.
For this, we define the MFG with the following separable reward,
\begin{equation}
    \rc(s,a,\rho) = \bar{r}(s,a) + \tilde{r}(s,\rho) \text{ with }
    \bar{r} = r \text{ and } \tilde{r}(\cdot,\rho) = \nabla G(\rho).
    \label{eq:reward_separable}
\end{equation}
We have the counterparts of Prop.~\ref{prop:general} and Thm.~\ref{thm:general}.
\begin{prop}
    \label{prop:special}
    The MFG defined Eq.~\eqref{eq:reward_separable} is (strictly)  monotone if and only if $G$ is (strictly) concave.
\end{prop}
\begin{proof}
    This is similar to the proof of Prop.~\ref{prop:general}. We have
    \begin{align}
        \text{$G$ concave} &\Leftrightarrow
        \forall \rho_1,\rho_2\in\Delta_\states,
        \langle \rho_1 - \rho_2, \nabla G(\rho_1) - \nabla G(\rho_2)\rangle \leq 0
        \\ &\Leftrightarrow
        \forall \rho_1,\rho_2\in\Delta_\states,
        \langle \rho_1 - \rho_2, \tilde{r}(\cdot,\rho_1) - \tilde{r}(\cdot,\rho_2)\rangle \leq 0.
    \end{align}
    The equivalence between the last condition and (strict) monotonicity follows from~\cite[Lemma~2]{perolat2021scaling}.
\end{proof}
\begin{theorem}
    \label{thm:special}
    Assume that $G$ is concave. Any maximizer of the CURL problem~\eqref{eq:curl_problem} with $F$ defined in Eq.~\eqref{eq:def_F_special} is a Nash equilibrium of the MFG defined through reward~\eqref{eq:reward_separable}, and conversely.
\end{theorem}
\begin{proof}
    The proof follows globally the same lines as the one of Thm.~\ref{thm:general}.
    As $G$ is concave in $\rho$, and $\rho$ being the marginal of $\mu$, $F$ is concave in $\mu$. Therefore, satisfying the optimality conditions provides a global optimizer. Before proceeding further, notice that $\nabla F(\mu) = r + \nabla_\mu G(\rho)$. Using the chain rule, we have
    \begin{equation}
        \frac{\partial G(\rho)}{\partial \mu(s,a)} = 
        \frac{\partial G(\rho)}{\partial \rho(s)} \frac{\partial \rho(s)}{\partial \mu(s,a)} =
        \frac{\partial G(\rho)}{\partial \rho(s)} \frac{\partial \sum_{a'\in\actions}\mu(s,a')}{\partial \mu(s,a)} =
        \frac{\partial G(\rho)}{\partial \rho(s)}.
    \end{equation}
    Therefore, we have
    \begin{align}
        \langle \mu', \nabla_\mu G(\rho)\rangle &=
        \sum_{s,a}\mu'(s,a) \frac{\partial G(\rho)}{\partial \rho(s)} =
        \sum_s \frac{\partial G(\rho)}{\partial \rho(s)} \sum_a \mu'(s,a) =
        \sum_s \frac{\partial G(\rho)}{\partial \rho(s)} \rho'(s)
        = \langle \rho', \nabla G(\rho)\rangle.
    \end{align}
    Hence, we have %
    \begin{flalign}
        \text{$\pi$ maximizer} 
        &\Leftrightarrow \langle \nabla F(\mu_\pi), \mu'-\mu_\pi\rangle \leq 0
        &\forall \mu'\in M
        \\
        &\Leftrightarrow \langle r + \nabla_\mu G(\rho_\pi), \mu' - \mu_\pi\rangle\leq 0
        &\forall \mu'\in M
        \\
        &\Leftrightarrow \langle \mu',r\rangle + \langle \rho',  \nabla G(\rho_\pi) \rangle \leq \langle \mu_\pi,r\rangle + \langle \rho_\pi,  \nabla G(\rho_\pi) \rangle
        &\forall \mu'\in M
        \\
        &\Leftrightarrow J(\pi',\mu_\pi) \leq J(\pi,\mu_\pi)
        &\forall \pi'\in\Delta_\actions^\states 
        \\
        &\Leftrightarrow \phi(\pi)=0 \Leftrightarrow \text{$\pi$ is Nash}.
    \end{flalign}
    This proves the stated result.
\end{proof}
One should be meticulous about uniqueness properties in this case. We have shown in Prop.~\ref{prop:special} the equivalence between the strict concavity of $G$ and the strict monotonicity of the game, hence uniqueness of the Nash. Yet, what is unique in this case is the population $\rho_\pi$, but not necessarily the joint distribution $\mu_\pi$ or the policy $\pi$
(\textit{e.g.}, an MDP with two actions having the same effect).

This equivalence between CURL and MFG is interesting, because any algorithm designed for
this setting of
MFGs can be readily applied to CURL, and conversely, at least when the MFG is potential and when the dynamics does not depend on the population.

\section{Algorithms}
\label{sec:algorithms}

Before presenting algorithms, we discuss how to measure their progresses. From a CURL viewpoint, it is natural to measure how $F$ increases with iterates. Yet, the maximum is usually not known a priori. From an MFG perspective, it may be more natural to use exploitability, %
that we know should be zero at optimality. Yet, it is more costly, as its evaluation requires computing a best response. %
Now, we will show that the seminal approach to CURL is indeed FP for MFGs, and connect it to Frank-Wolfe. Then, we will discuss briefly other algorithms specific to either CURL or MFGs.

\subsection{On \citet{hazan2019provably} Approach, Fictitious Play and Frank-Wolfe}

\begin{algorithm}
    \caption{\citet{hazan2019provably} algorithm.\label{alg:hazan}}
    \KwIn{Step size $\eta$, number of iterations $T$, initial policy $\pi_0$}
    Set $C_0=(\pi_0)$, $\alpha_0=1$, $\pi_{\text{mix},0} = (\alpha_0, C_0)$\;
    \For{$t=0,\dots,T-1$}{
        Compute $\pi_{t+1}$, optimal for reward $\nabla G(\rho_{\pi_{\text{mix},t}})\in\R^\states$\;
        Update $\pi_{\text{mix},t+1} = (\alpha_{t+1}, C_{t+1})$ with
        $C_{t+1} = (C_t,\pi_{t+1})$ and $\alpha_{t+1} =((1-\eta)\alpha_t, \eta)$\;
    }
    \Return{$\pi_{\text{mix},T}$}
\end{algorithm}

\textbf{\citet{hazan2019provably} approach.}
As far as we know, the first approach that addresses the general CURL problem was proposed by~\citet{hazan2019provably}. They consider the case $F(\mu) = G(\rho)$ (so a special case of what was studied in Sec.~\ref{sec:curl_as_mfg}, but the approach could readily be extended to a general $F(\mu)$). For a set of $k$ policies $C = (\pi_0,\dots,\pi_{k-1})$ and weights $\alpha\in\Delta_{\{0,\dots,k-1\}}$, they define the non-stationary policy $\pi_\text{mix}$ as the one that samples a policy $\pi_i$ with probability $\alpha_i$ in the initial state, and this policy is used for the whole trajectory. The induced occupancy measure satisfies $\rho_{\pi_\text{mix}} = \sum_{i=0}^{k-1} \alpha_i \rho_{\pi_i}$.
Their approach is inspired by Frank-Wolfe~\citep{frank1956algorithm}, and is depicted in Alg.~\ref{alg:hazan} (here assuming a known model). The mixing policy is initialized with some given policy. At  iteration $t$, a new policy being the best response to the MDP defined through the reward $\nabla G(\rho_{\pi_{\text{mix},t}})$ is added to the mixture, and the set of weights is updated according to a predefined learning rate. Therefore, the CURL problem is reduced to a sequence of MDP problems.
Notice that what characterizes the optimality of the policy is the state occupancy measure it induces. Therefore, instead of returning the non-stationary policy $\pi_{\text{mix},T}$, one could return the stationary policy
$\pi_{\text{sta},T}(a|s) = {\sum_{t=0}^T\alpha_{T}(t) \rho_{\pi_t}(s)\pi_t(a|s)}({\sum_{t=0}^T \alpha_T(t)\rho_{\pi_t}(s)})^{-1}$, as it induces the same state occupancy measure. This implies that Alg.~\ref{alg:hazan} is indeed Fictitious Play, as explained further later.
Notice also that on an abstract way, the approach of~\citet{lee2019efficient} is exactly the same: it differs from~\citep{hazan2019provably} in the kind of function $G$ considered (specifically a KL divergence for~\citep{lee2019efficient}), and on how approximations are done without knowledge of the domain (something we will not address here, but that comes with guarantees for~\citet{hazan2019provably}).

\begin{algorithm}%
    \caption{Fictitious Play~\citep{perrin2020fictitious}.\label{alg:fp}}
    \KwIn{number of iterations $T$, initial policy $\pi_0$}
    Set $\bar{\mu}_0 = \mu_{\pi_0}$\;
    \For{$t=0,\dots,T-1$}{
        Compute $\pi_{t+1} \in \argmax_{\pi\in\Delta_\actions^\states} J(\pi, \bar{\mu}_t)$\;
        Update $\bar{\mu}_{t+1} = \frac{1}{t+2} \mu_{\pi_{t+1}} + \frac{t+1}{t+2} \bar{\mu}_t$\;
    }
    \Return{$\bar{\pi}_{T}: \bar{\pi}_T(a|s) = \frac{\bar{\mu}_T(s,a)}{\bar{\rho}_T(s)}$}
\end{algorithm}%

\textbf{Fictitious Play.}
It is originally an iterative algorithm for repeated games~\citep{robinson1951iterative}, where each agent plays optimally according to an empirical expectation of past observed strategies. It has been adapted to MFGs~\citep{cardaliaguet2017learning,perrin2020fictitious} with the approach summarized in Alg.~\ref{alg:fp}. This is for the case where $\rc\in\R^{\states\times\actions\times\Delta_{\states\times\actions}}$, but one just needs to replace $J(\pi,\bar{\mu}_t)$ by $J(\pi,\bar{\rho}_t)$ (with $\bar{\rho}$ still the marginal of $\bar{\mu}$, as defined Eq.~\eqref{eq:def_rho}) to cover the other case. The algorithm is initialized for some $\pi_0$, and at each iteration a best response is computed by maximizing $J(\pi,\bar{\mu}_t)$ (Eq.~\eqref{eq:mfg_criterion}), with $\bar{\mu}_t$ the average of all past state-action occupancy measures (associated to all past best responses). The output policy is the stationary one having $\bar{\mu}_T$ as associated occupancy measure (see Sec.~\ref{sec:background}).
With this, we can easily see  that Alg.~\ref{alg:hazan} is indeed an FP approach, equivalent to Alg.~\ref{alg:fp}. With the choice $\eta = \frac{1}{t+2}$, we have that $\rho_{\pi_{\text{mix},t}} = \bar{\rho}_t$. With the MFG defined in Sec.~\ref{sec:curl_as_mfg}, that is, with reward $\rc(\cdot,\rho) = \nabla G(\rho)$ (Eq.~\eqref{eq:reward_separable}), maximizing $J(\pi,\bar{\rho}_t)$ is equivalent to solving the MDP with reward $\nabla G(\rho_{\text{mix},t})$. Eventually, both policies $\pi_{\text{mix}, T}$ (non-stationary) and $\bar{\pi}_T$ (stationary) induce the same state occupancy measure, and thus the same solution to the problem. 

\textbf{Frank-Wolfe.}
Now, we take advantage of the relationship between CURL and MFGs to provide a new interpretation of FP for (potential) MFGs as equivalent to Frank-Wolfe (FW) applied to the original optimization problem. This notably allows to provide the first convergence rate for discrete-time FP in this setting.
As seen before, CURL~\eqref{eq:curl_problem} can be defined solely in terms of occupancy measure, $\max_{\mu\in M} F(\mu)$. The set $M$~\eqref{eq:def_M} being defined as a set of linear constraints, it is a convex set. With $F$ concave and $M$ convex, we can readily apply FW. With $\mu_0\in M$, for all $t\geq 0$,
\begin{equation}
    \mu_{t+1/2} \in \argmax_{\mu\in M} \langle \mu, \nabla F(\mu_t)\rangle
    \text{ and }
    \mu_{t+1} = (1-\eta_{t+1})\mu_t + \eta_{t+1} \mu_{t+1/2}.
\end{equation}
First, one searches for the element of $M$ which is the most collinear with the gradient. Then, this is used to update the estimate, so as to stay in the domain, $\mu_{t+1} = (1-\eta_{t+1})\mu_t + \eta_{t+1} \mu_{t+1/2}$. With $\eta_t = \eta$, we retrieve the occupancy measures induced by the mixing policy of~\cite{hazan2019provably}, and with $\eta_t = \frac{1}{t+1}$ we obtain the averaging of FP.  Now, for the inner optimization problem, observe that
\begin{equation}
    \max_{\mu\in M} \langle \mu, \nabla F(\mu_t)\rangle = \max_{\pi\in\Delta_\actions^\states} \langle \mu_\pi, \nabla F(\mu_t)\rangle,
\end{equation}
which corresponds exactly to the best response computed in both Algs.~\ref{alg:hazan} and~\ref{alg:fp}. So, both are indeed special instances of FW. This is not new for the approach of~\citet{hazan2019provably}, who refer explicitly to FW (but do not make any connection to FP or MFGs, or games in general), but this is new regarding FP for (potential) MFGs, as far as we know.
This leads to the following result.
\begin{theorem}
    \label{thm:frank_wolfe}
    Consider a  monotone potential MFG, that is such that $\rc(\cdot,\mu) = \nabla F(\mu)$ with $F$ concave (Prop.~\ref{prop:general}). Write $R=\max_{\mu,\mu'\in M}\|\mu-\mu'\|$ the diameter of $M$. Let $\mu_*\in M$ be a maximizer of $F$ (or equivalently by Thm.~\ref{thm:general}, let 
    $\pi_*(a|s) = \mu_*(s,a)/\rho_*(s)$ be a Nash of the MFG). Assume that the reward is $\beta$-Lipschitz in the sense that
    $%
        \forall \mu,\mu'\in\Delta_{\states\times\actions}, \; \|\rc(\cdot,\mu) - \rc(\cdot,\mu')\| \leq \beta \|\mu - \mu'\|%
    $. %
    Generally speaking, the exploitability is bounded by the suboptimality. For any $\pi\in\Delta_\actions^\states$:
    \begin{equation}
        \phi(\pi) \leq F(\mu_*) - F(\mu_\pi) + R\sqrt{2\beta(F(\mu_*) - F(\mu_\pi))}.
    \end{equation}
    Specifically, by running FW/FP for $T$ iterations, with $\eta_t = \frac{2}{t+1}$,
    \begin{equation}
        \phi(\pi_T) \leq \frac{2\beta R^2}{\sqrt{T+1}} + \frac{2\beta R^2}{T+1} \text{ with } \pi_T(a|s) = \frac{\mu_T(s,a)}{\rho_T(s)}.
    \end{equation}
\end{theorem}
\begin{proof}
    First, notice that the $\beta$-Lipschitzness of the reward is equivalent to the classic $\beta$-smoothness of the function $F$. 
    For the first result, we will exploit the equivalence between upper-bounding the exploitability $\phi(\pi)$ and upper-bounding the optimality condition $\langle \nabla F(\mu_\pi), \mu' - \mu_\pi\rangle$ (for arbitrary $\mu'\in M$). Indeed, from the proof of Thm.~\ref{thm:general}, let $\epsilon>0$ and $\pi\in\Delta_\actions^\states$ be an arbitrary policy,
    \begin{equation}
        \forall \mu'\in M, \langle \nabla F(\mu_\pi),\mu'-\mu_\pi\rangle \leq \epsilon 
        \Leftrightarrow
        \phi(\pi) \leq \epsilon.
        \label{eq:equivalence}
    \end{equation}
    For any $\mu'\in M$, with a simple decomposition, we have that
    \begin{equation}
        \langle \nabla F(\mu_\pi), \mu' - \mu_\pi\rangle = 
        \langle \nabla F(\mu_*), \mu' - \mu_\pi\rangle
        + \langle \nabla F(\mu_\pi) - \nabla F(\mu_*) , \mu' - \mu_\pi\rangle.
    \end{equation}
    We now upper-bound each of these two terms. For the first one:%
    \begin{align}
        \langle \nabla F(\mu_*), \mu' - \mu_\pi\rangle
        &= \langle \nabla F(\mu_*), \mu' - \mu_*\rangle
        + \langle \nabla F(\mu_*), \mu_* - \mu_\pi\rangle 
        \\
        &\myeq{a}{\leq} \langle \nabla F(\mu_*), \mu_* - \mu_\pi\rangle
        \\
        &\myeq{b}{\leq} F(\mu_*) - F(\mu_\pi),
    \end{align}
    with \emph{(a)} by optimality conditions and \emph{(b)} by concavity of $F$.
    For the second term, we have the following, using notably a classic result from convex optimization for $\beta$-smooth functions,
    \begin{align}
        \langle \nabla F(\mu_\pi) - \nabla F(\mu_*) , \mu' - \mu_\pi\rangle
        &\myeq{a}{\leq} \|\nabla F(\mu_\pi)-\nabla F(\mu_*)\| \|\mu'-\mu_\pi\| %
        \\
        &\myeq{b}{\leq} R \|\nabla F(\mu_\pi)-\nabla F(\mu_*)\| 
        \\
        &\myeq{c}{\leq} R \sqrt{2\beta \left(\langle \nabla F(\mu_*), \mu_\pi - \mu_*\rangle + F(\mu_*) - F(\mu_\pi) \right)}
        \\
        &\myeq{d}{\leq} R\sqrt{2\beta (F(\mu_*) - F(\mu_{\pi}))},
    \end{align}
    with \emph{(a)} by Cauchy-Schwartz, \emph{(b)} by def. of the diameter, \emph{(c)} by \cite[Lemma~3.5]{bubeck2015convex} applied to $-F$ and \emph{(d)} by optimality conditions.
    Combining both bounds with the equivalence result of Eq.~\eqref{eq:equivalence}, we get the first stated result.
    For the second inequality, the set $M$ being convex, and $F$ being concave, we can apply the classic convergence result for FW (\textit{e.g.},~\cite[Thm.~3.8]{bubeck2015convex}) to obtain
    \begin{equation}
        F(\mu_*) - F(\mu_T) \leq \frac{2\beta R^2}{T+1}.
    \end{equation}
    Plugging this into the exploitability bound allows concluding.
\end{proof}

\looseness=-1
\citet{perrin2020fictitious}  showed that continuous-time FP (time of iterations, so a theoretical construct) enjoys a rate $\phi(T) = O(T^{-1})$ for  monotone MFGs. They conjectured a rate $\phi(T) = O(T^{-1/2})$ for discrete-time FP. This is exactly what provides Thm.~\ref{thm:frank_wolfe}, yet under stronger assumptions (the MFG needs to be potential, and the reward $\beta$-Lipschitz). The reward smoothness may not be always satisfied, as for the entropy example. %
Yet, a smoothed objective can also be considered (\textit{e.g.}, a smoothed version $\hc_\epsilon(\rho) = -\langle \rho, \ln(\rho+\epsilon)\rangle$ of the entropy~\citep{hazan2019provably}). %
Notice also that we prove that the exploitability is upper-bounded by the suboptimality. Thus, any better bound on FW (\textit{e.g.}, based on stronger assumptions) would translate readily into a faster decrease of the exploitability.
FW has been extended to saddle-point problems, and in the bilinear case shown to be equivalent to FP~\cite{hammond1984solving,gidel2017frank}. However, our connection between FW and FP for MFGs is new, to the best of our knowledge.

\subsection{Online Mirror Descent}
\label{subsec:omd}

\begin{algorithm}%
    \caption{Online Mirror Descent \citep{perolat2021scaling}.\label{alg:omd}}
    \KwIn{number of iterations $T$, initial policy $\pi_0$, parameter $\alpha$}
    Set $Q_0 = 0$\;
    \For{$t=0,\dots,T-1$}{
        Compute $q_{\pi_t, \rc(\cdot,\mu_{\pi_t})}$\;
        Update $Q_{t+1} = Q_t + \alpha q_{\pi_t, \rc(\cdot,\mu_{\pi_t})}$\;
        Define $\pi_{t+1} = \sm(Q_{t+1})$\;
    }
    \Return{$\pi_T$}
\end{algorithm}%

To emphasize  the potential benefits of addressing the CURL problem under the MFG point of view, we specifically consider the Online Mirror Descent algorithm (Alg.~\ref{alg:omd}), recently adapted to solve MFGs~\citep{perolat2021scaling}. It is empirically faster than FP on various MFGs and comes with additional advantages. The algorithm is initialized with a policy $\pi_0$. At each iteration, the policy $\pi_t$ being given, one can compute or estimate the associated occupancy measure $\mu_{\pi_t}$ (as before, considering a state occupancy measure just amounts to replacing $\mu$ by $\rho$). This being fixed, one can compute the state-action value function $q_{\pi_t,\rc(\cdot,\mu_{\pi_t})}\in\R^{\states\times\actions}$ of the policy $\pi_t$ with the reward being $r=\rc(\cdot,\mu_{\pi_t})$. This is a first advantage compared to FP: one simply needs to perform a policy evaluation step instead of computing a best response (that is, solving a full MDP). Then, the scaled value function is accumulated in $Q_{t+1}\in\R^{\states\times\actions}$. This is a second advantage compared to FP: it might be easier to accumulate value functions rather than occupancy measures, especially in an approximate case (not covered here, again). Eventually, the policy is simply the softmax of the scaled sum of past value functions (hence the parameter $\alpha$ can be understood as an inverse temperature). \citet{perolat2021scaling} show the convergence of a continuous-time variation of this algorithm (here, the time is the iteration time $t$, so this corresponds to a theoretical abstraction of Alg.~\ref{alg:omd}).
Notice that this OMD for MFGs can be seen as a generalization of Mirror-Descent Policy Iteration~\citep{geist2019theory} (retrieved for linear $F$), which is an abstraction of efficient deep RL algorithms such as TRPO~\citep{schulman2015trust}, MPO~\citep{abdolmaleki2018maximum} or M-RL~\citep{vieillard2020munchausen}, and which comes with strong theoretical guarantees in the RL setting~\citep{vieillard2020leverage}. Thus, provided a good density estimator (\textit{e.g.}, a normalizing flow~\citep{papamakarios2021normalizing}), this could be a good basis for a deep MFG agent, but this is beyond the scope of the current paper.

\subsection{Other Algorithms}
\label{subsec:otherAlgs}

\looseness=-1
\textbf{CURL-related algorithms. } Apart from~\cite{hazan2019provably} discussed above,  \citet{zhang2020variational} also address the general problem, by introducing a policy gradient approach to CURL. It is a gradient ascent of $F(\mu_\pi)$ according to $\pi$. Even though $F$ is not concave in $\pi$, they provide a global convergence result for their approach, exploiting a so-called ``hidden concavity'' (roughly speaking, exploiting the fact that $\Delta_\actions^\states$ and $M$ are in bijection, which is true at least for the tabular case). The naive gradient $\nabla_\pi F(\mu_\pi)$ (obtained using the chain-rule) cannot be estimated in general. They thus exploit the Legendre-Fenchel transform of $-F$ to express this gradient as the solution of a saddle-point problem, involving quantities that are easier to estimate. As such, their approach can be considered as a reduction from CURL to a sequence of (zero-sum) games. We will focus for the rest of this paper on the simpler case where CURL is reduced to a sequence of MDP problems, at most.
The concurrent work of~\citet{zahavy2021reward} provides a meta-algorithm for solving the CURL problem, that can be instantiated by combining regret minimizing algorithms (one for the best response, one for computing the reward as a function of all past occupancy measures). They also retrieve FW as a special case. Contrary to OMD, they require to compute best responses (except if relying on non-stationary, finite horizon RL), and the reward to be considered at each iteration depends on all past occupancy measures (instead of the last one only).
It also relies on the conjugate of $F$.

\textbf{MFG-related algorithms. }
We covered the MFG literature in Sec.~\ref{sec:background}, especially the part dealing with MDPs and learning. However, many of these approaches, \textit{e.g.}~\cite{subramanian2019reinforcement,guo2019learning,anahtarci2019fitted,anahtarci2020q,xie2020provable}, rely on a fixed-point setting, and make an implicit ergodic assumption, effectively getting ride of the initial state distribution. We explain briefly why, ignoring the additional dependency of the dynamics to the population usually considered. For $\rho\in\Delta_\states$, define $\Gamma_1(\rho)\in\Delta_\actions^\states$ the optimal policy for the reward $\rc(\cdot,\rho)$. Slightly abusing notations, write $P_\pi$ the policy-induced state-transition kernel ($P_\pi(s'|s) = \sum_a \pi(a|s) P(s'|s,a)$). For $\pi\in\Delta^\states_\actions$ and $\rho\in\Delta_\states$, define $\Gamma_2(\pi, s) = P_\pi^\top \rho \in\Delta_\states$, the state distribution obtained by starting from $\rho$ and applying $\pi$ for one step. Then, define $\Gamma(\rho) = \Gamma_2(\Gamma_1(\rho),\rho)\in\Delta_\states$. These works make structural assumptions about the MDP so that $\Gamma$ is a contraction (which do not hold for finite MDPs
~\cite{cui2021approximately}), and the Nash equilibrium is then the fixed point of this operator, $\rho_* = \Gamma(\rho_*)$. First, notice that $\rho^*$ is a stationary distribution, not a discounted occupancy measure, so $\rho^*\notin M$. So, this does not provide a solution to the defined CURL problem. Second, this notion of Nash equilibrium is different from the one we considered. We do require a representative player to not be able to gain something when both this player and the population start from the initial state distribution. With the fixed-point approach, it is required from the representative player to not be able to gain something when the population is already at equilibrium (that is, has reached its asymptotic behavior). For these reasons, applying the family of related algorithms to the CURL problem is not obvious\footnote{Alternatively, we could define a variant of the CURL problem with stationary rather than discounted occupancy measures, and this family of algorithms could apply.}. Yet, algorithms framed for the setting considered here (and not introduced in this paper) can already be of interest for CURL, as illustrated empirically next. 

\section{Numerical Illustration}

\looseness=-1
This section aims at providing additional evidences of the benefits from connecting CURL and MFGs, %
through a set of numerical illustrations. We consider an environment simple enough to be visualized and instantiate on it a number of special cases of CURL, depicted in Sec.~\ref{subsec:curl}. We consider the model to be known and compute  best responses exactly (or do exact policy evaluation). Our aim is to compare a representative algorithm for CURL, the one from~\citet{hazan2019provably} that we have shown to be equivalent to FP in an MFG context, to an algorithm only considered in the MFG setting, namely OMD~\citep{perolat2021scaling}. What we do not address here is the approximate case. Yet, each of the problems considered  is not straightforward, even with a known model. Moreover, we think that density estimation methods considered in the literature~\citep{hazan2019provably,lee2019efficient} for \citet{hazan2019provably} approach could be applied to OMD (see also discussion Sec.~\ref{subsec:omd}).

\begin{figure}[tbh]
    \centering
    \begin{minipage}{.19\linewidth}
    \includegraphics[width=.85\linewidth]{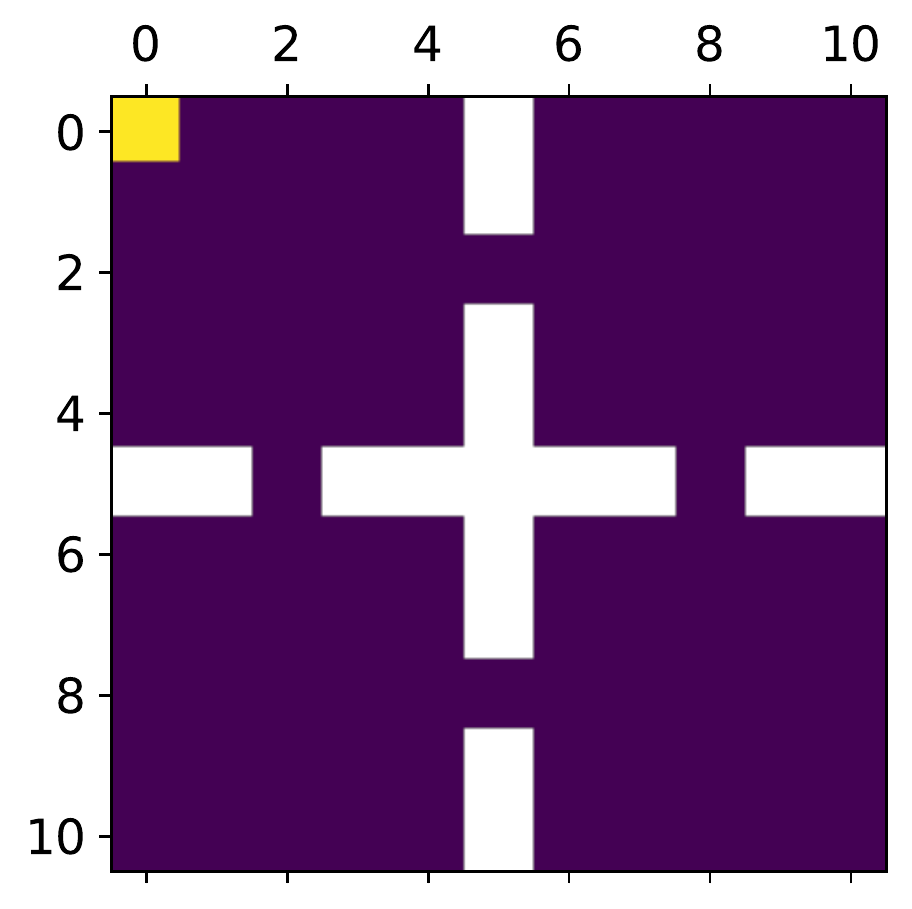}
    \end{minipage}
    \begin{minipage}{.19\linewidth}
    \includegraphics[width=\linewidth]{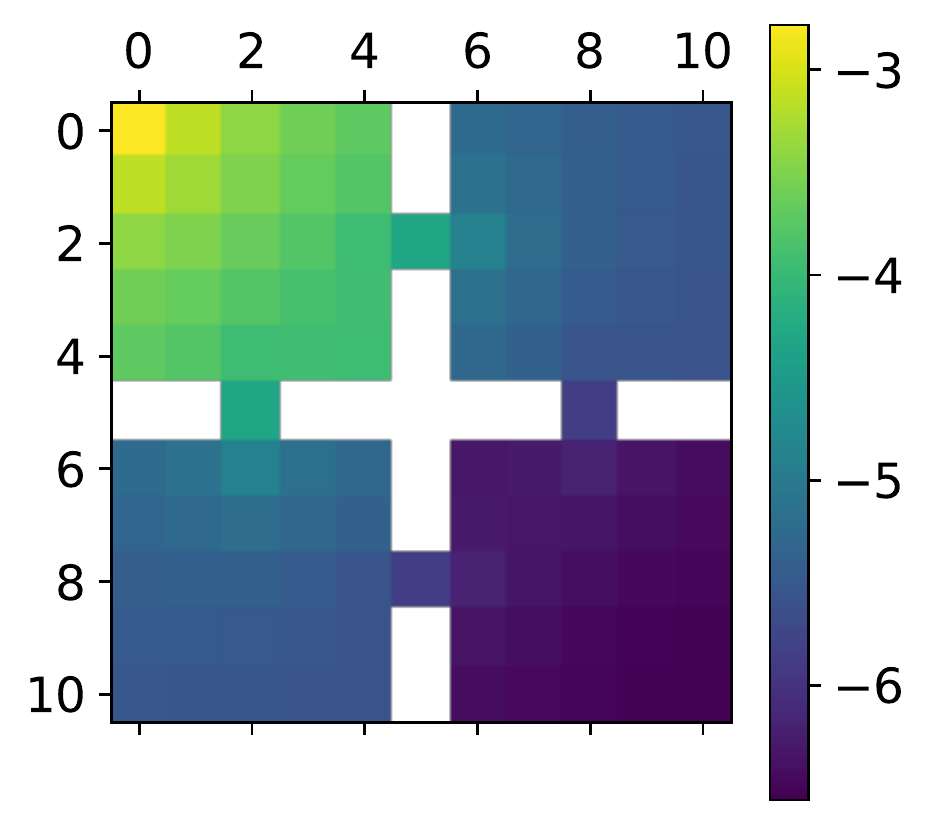}
    \end{minipage}
    \begin{minipage}{.19\linewidth}
    \includegraphics[width=\linewidth]{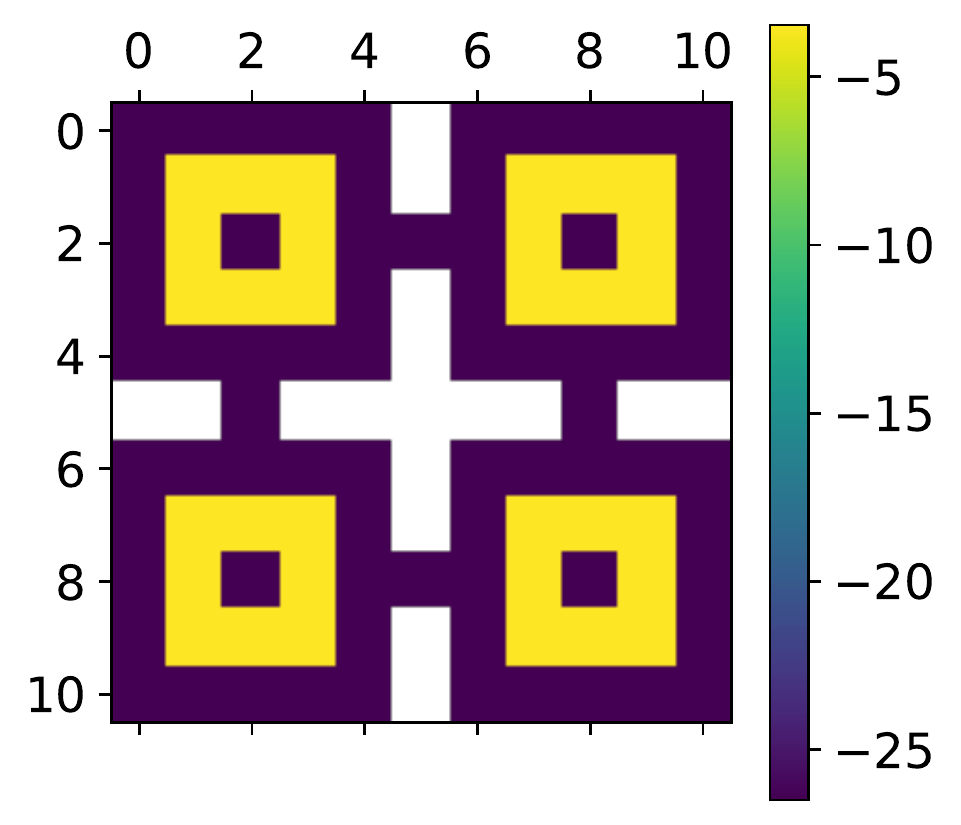}
    \end{minipage}
    \begin{minipage}{.19\linewidth}
    \includegraphics[width=.85\linewidth]{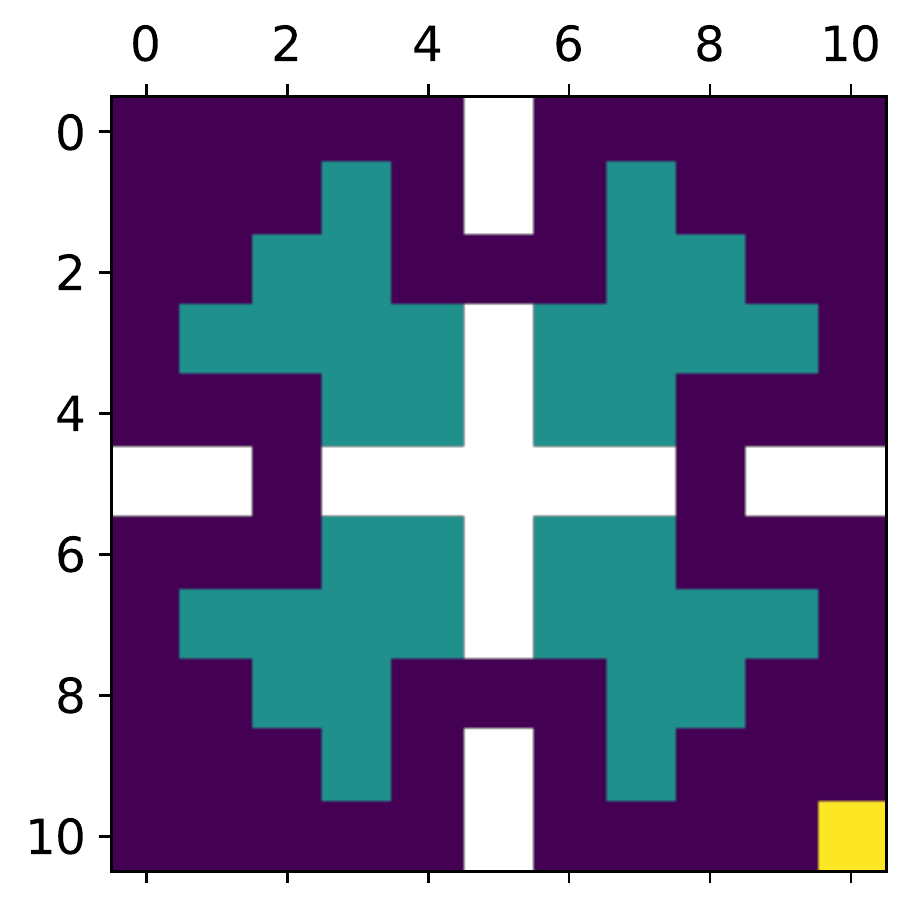}
    \end{minipage}
    \begin{minipage}{.19\linewidth}
    \includegraphics[width=.85\linewidth]{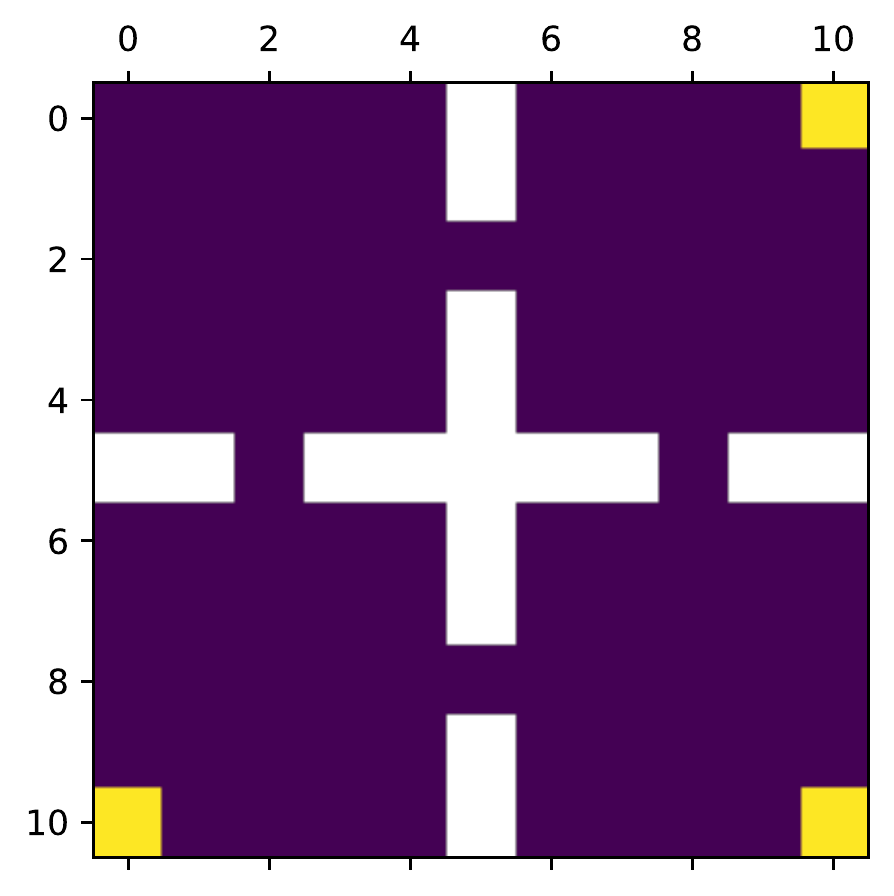}
    \end{minipage}
    \caption{%
    Reading order: \textbf{(a)} the considered environment initial state in yellow, walls in white); \textbf{(b)} the log-density of a uniform policy (to illustrate entropy maximization); \textbf{(c)} the target log-density for marginal matching; \textbf{(d)} the constrained MDP (reward in yellow,  constraints in light blue); \textbf{(e)} the three rewards of multi-objective RL. 
    }
    \label{fig:settings}
\end{figure}

Fig.~\ref{fig:settings} shows the considered environment and setting. The environment is a simple four-rooms problem (four actions, up, down, left and right; moving towards a wall does not change the state; deterministic dynamics), the initial distribution being a Dirac on the upper-left corner. We consider four settings: entropy maximization (the log-density of the population corresponding to a uniform policy being given in Fig.~\ref{fig:settings} as a baseline), marginal matching (using a KL-divergence, the target distribution being in Fig.~\ref{fig:settings}; notice that it is not achievable by an agent, the support being not connected), a constrained MDP (cost and reward are depicted in Fig.~\ref{fig:settings}), and multi-objective RL (with three rewards depicted Fig.
~\ref{fig:settings}). For each problem, we compare FP (that is, the seminal approach of~\cite{hazan2019provably} to CURL) to OMD (specific to MFGs so far), both regarding the optimized function (CURL viewpoint) and the exploitability (MFG viewpoint), as a function of the number of policy evaluations (FP is run for 300 iterations, and then OMD is run to have approximately the same number of policy evaluations). We also show the population (that is, occupancy measure) computed by each approach. %

For FP, we used the  FP rate $\eta_{t} = \frac{1}{t+1}$. We also tried FW rate ($\eta_t = \frac{2}{t+1}$), which provides similar results, and a constant rate~\citep{hazan2019provably} ($\eta_t = \eta$), which depends on the value of $\eta$ and is never better than FP or FW. This is to be expected, this constant rate was designed to offer guarantees in the approximate case, while we consider the exact one. For OMD, in all experiments, we use $\alpha=0.01$.

\begin{figure}[tbh]
    \centering
    \begin{minipage}{.3\linewidth}
    \includegraphics[width=\linewidth]{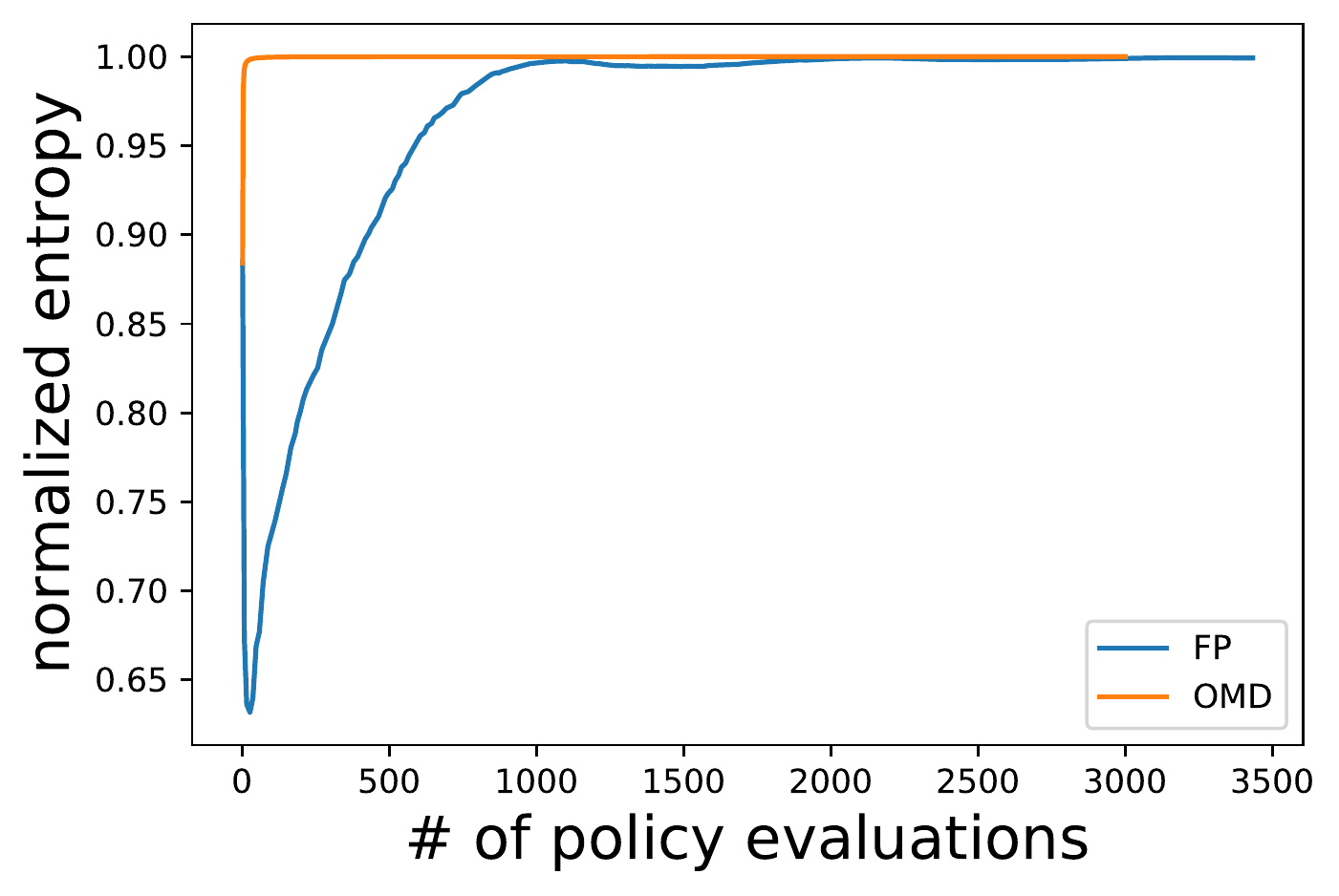}
    \end{minipage}
    \begin{minipage}{.3\linewidth}
    \includegraphics[width=\linewidth]{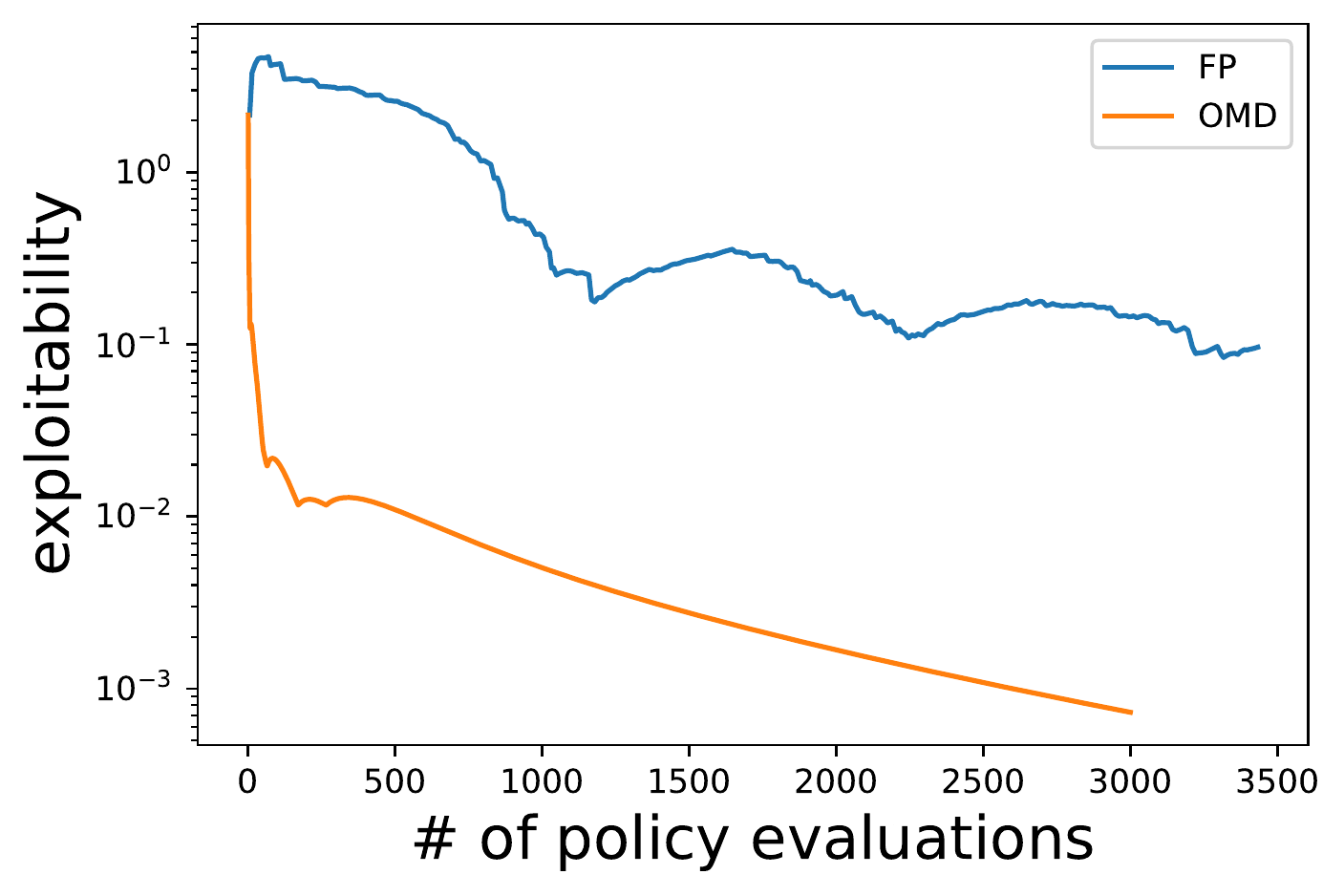}
    \end{minipage}
    \begin{minipage}{.19\linewidth}
    \includegraphics[width=\linewidth]{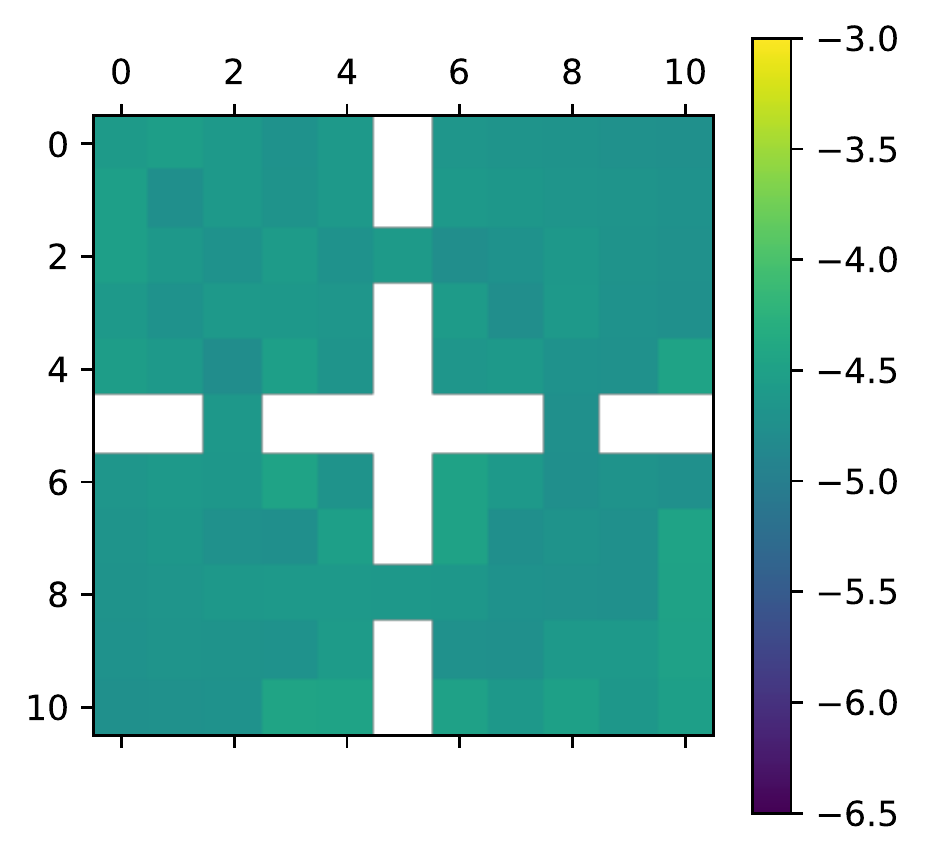}
    \end{minipage}
    \begin{minipage}{.19\linewidth}
    \includegraphics[width=\linewidth]{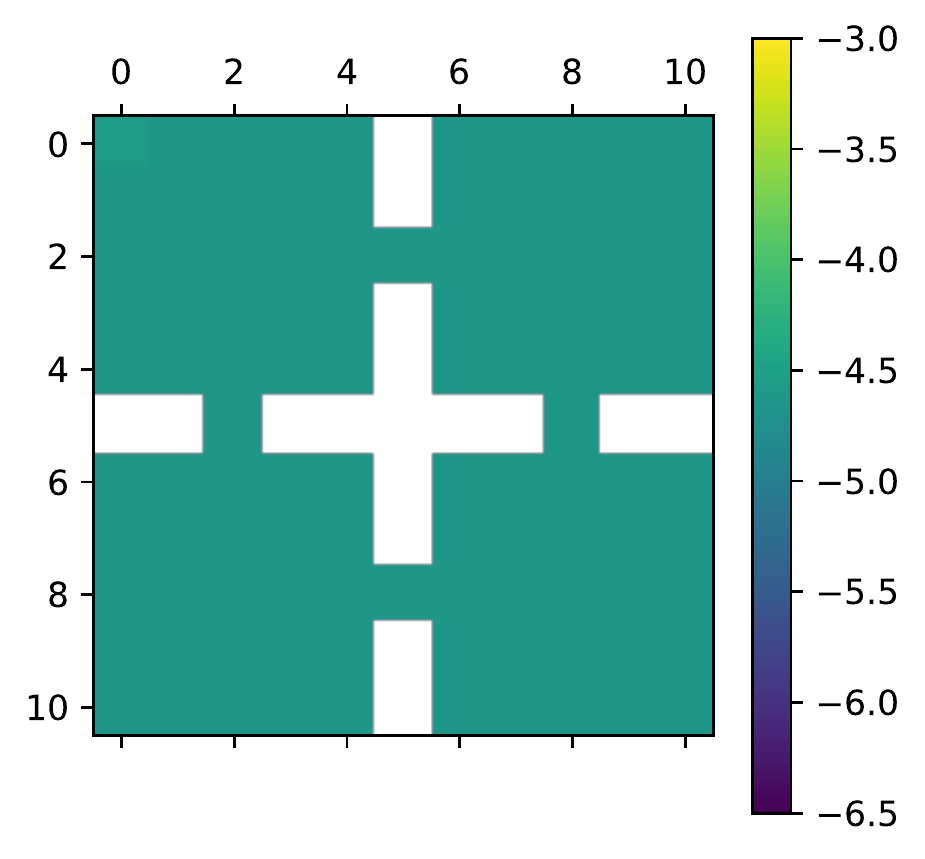}
    \end{minipage}
    \caption{\textbf{Entropy maximization.} 
    Reading order: \textbf{(a)} normalized entropy; \textbf{(b)} exploitability; \textbf{(c)}~log-density for the FP population; \textbf{(d)} log-density for the OMD population. 
    }
    \label{fig:entropy}
\end{figure}

\textbf{Entropy maximization.} 
The optimized function is $ F_\text{ent}(\mu) = G_\text{ent}(\rho) = -\langle \rho, \ln\rho\rangle$, with related reward $\rc_\text{ent}(\cdot,\mu) = \nabla G_\text{ent}(\rho) = - \ln \rho -1$. 
Results are presented Fig.~\ref{fig:entropy} (the entropy is normalized by $\ln(|\states|)$, the entropy of a uniform distribution). We can observe that OMD is much faster than FP both at increasing the entropy and decreasing the exploitability. At the final iteration, the entropies are hardly distinguishable numerically, but one can observe on the log-density plots that the population of OMD is more uniform (the scale is the same as for the uniform policy in Fig.~\ref{fig:settings}.b, to ease comparison), which is consistent with the much lower exploitability.

\begin{figure}[tbh]
    \centering
    \begin{minipage}{.3\linewidth}
    \includegraphics[width=\linewidth]{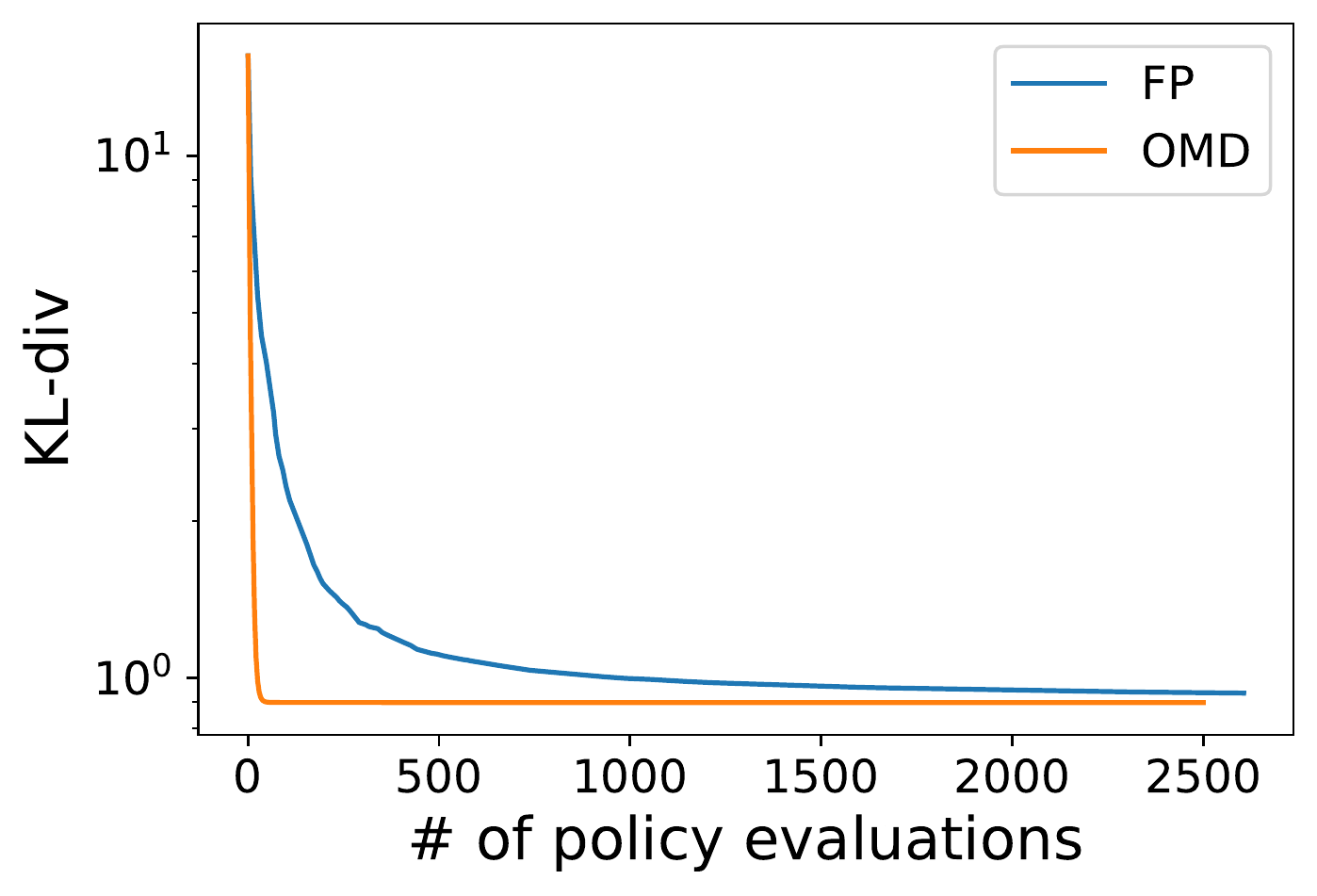}
    \end{minipage}
    \begin{minipage}{.3\linewidth}
    \includegraphics[width=\linewidth]{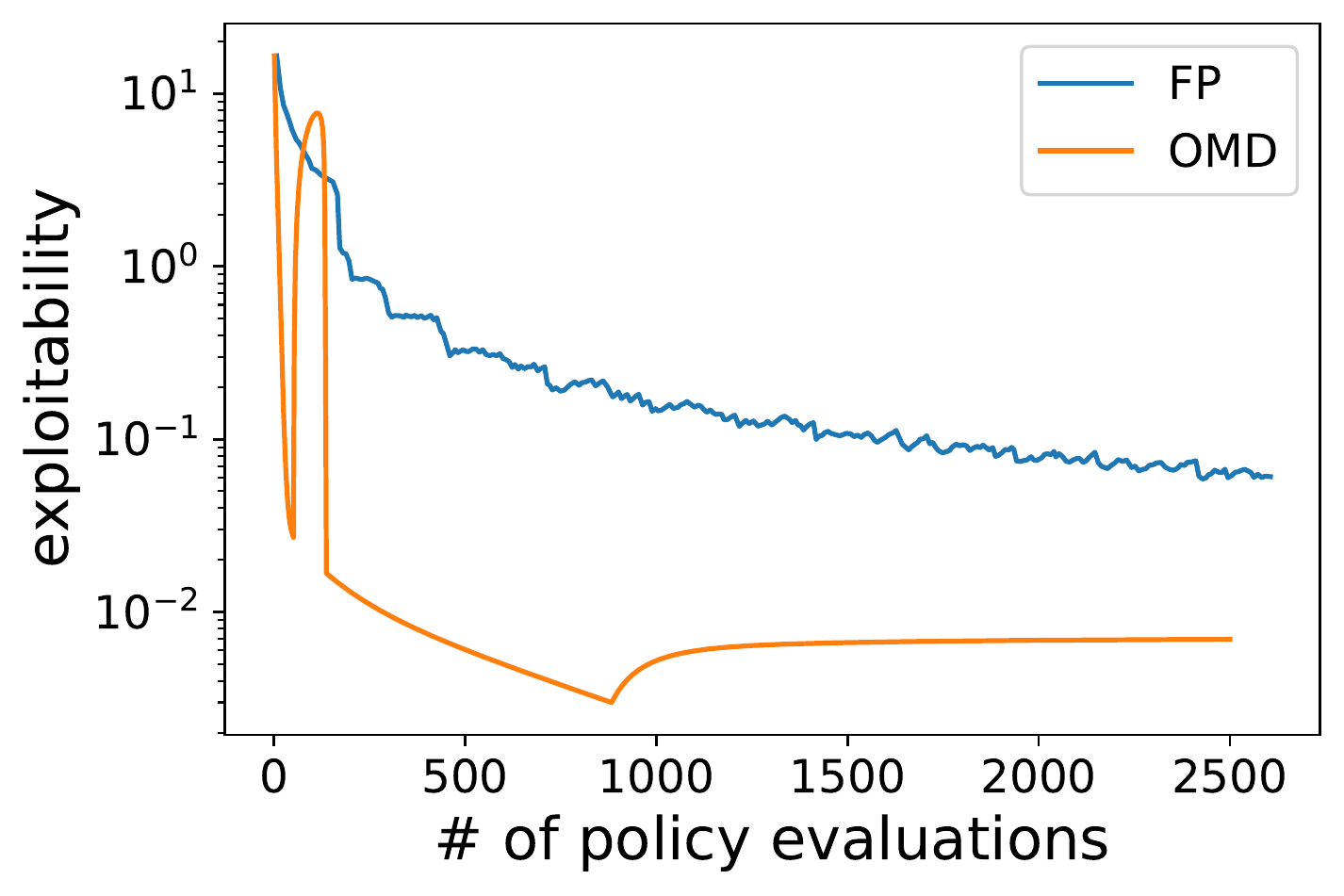}
    \end{minipage}
    \begin{minipage}{.19\linewidth}
    \includegraphics[width=\linewidth]{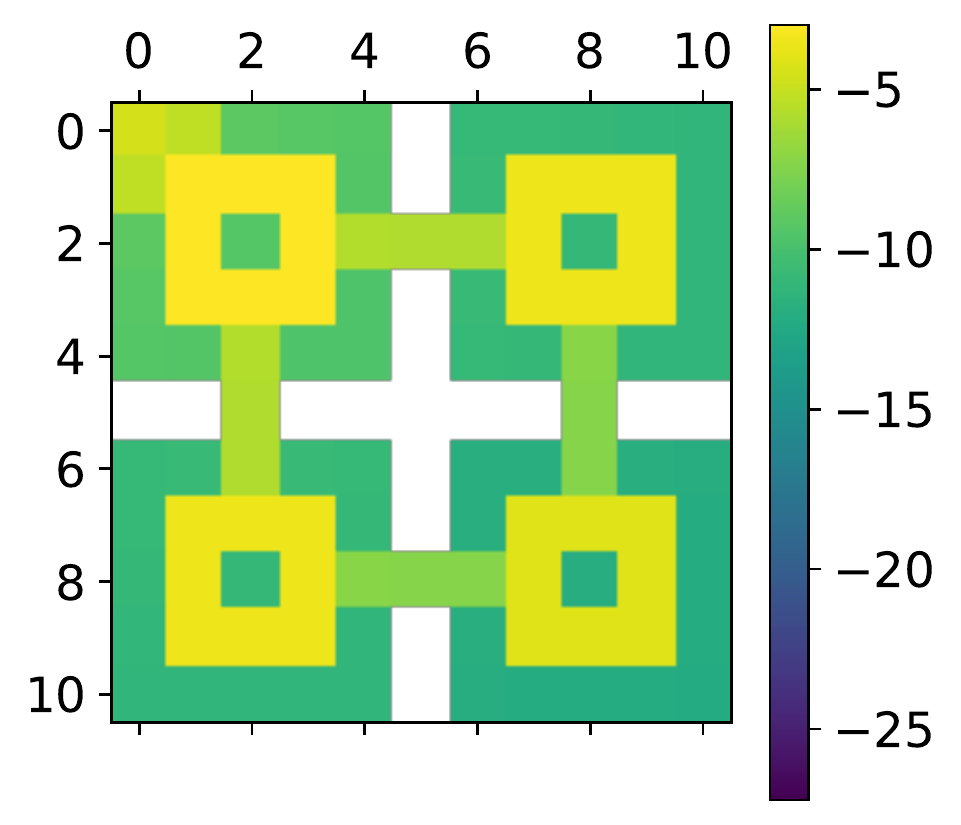}
    \end{minipage}
    \begin{minipage}{.19\linewidth}
    \includegraphics[width=\linewidth]{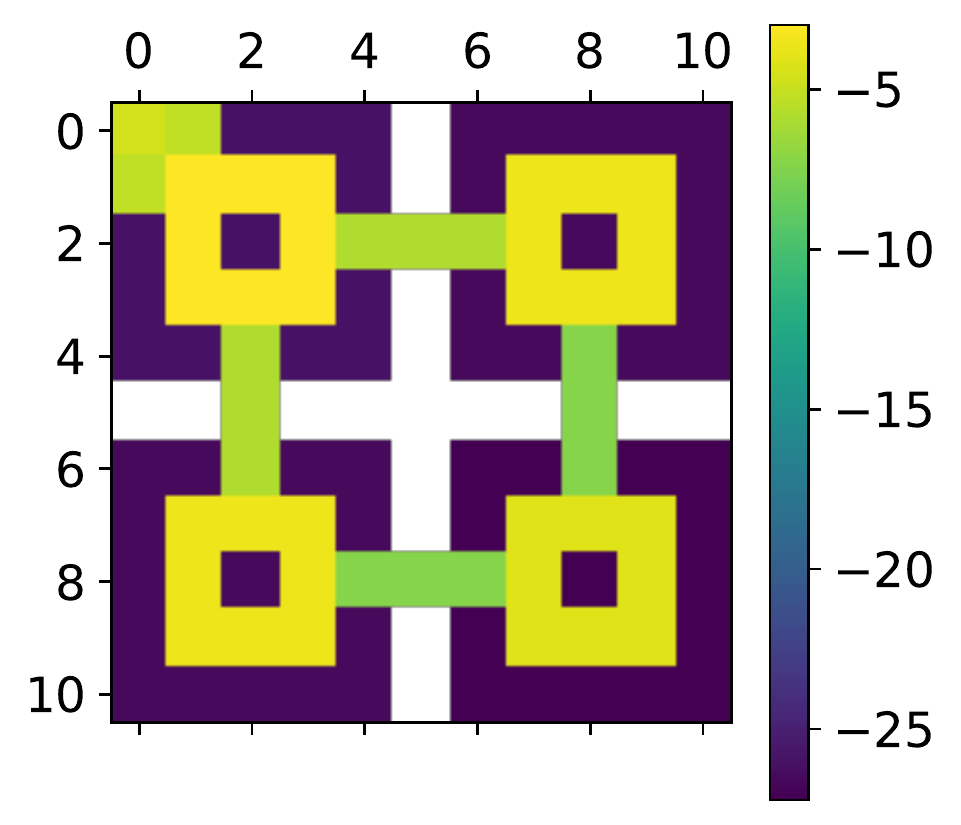}
    \end{minipage}
    \caption{\textbf{Marginal matching.} 
    Reading order: \textbf{(a)} divergence; \textbf{(b)} exploitability; \textbf{(c)}~log-density for the FP population; \textbf{(d)} log-density for the OMD population.
    }
    \label{fig:mm}
\end{figure}

\textbf{Marginal matching.}
Here, the optimized function is $F_\text{mm}(\mu) = G_\text{mm}(\rho) = -\kl{\rho}{\rho_*}$, with $\rho_*$ depicted in Fig.~\ref{fig:settings}, and the corresponding reward is $\rc_\text{mm}(\cdot,\mu) = \nabla G_\text{mm}(\rho) = \ln\rho_* - \ln \rho -1$. 
Results are presented Fig.~\ref{fig:mm}. Purposely, the target density is chosen such that it is not reachable by an agent. This illustrates the interest of the exploitability, which is zero at optimality, while we don't know a priori what the best solution is in terms of divergence. Here again, OMD is much faster at both optimizing the objective function and decreasing the exploitability (it is not monotonic for OMD, but nothing says it should be). We also observe from the log-density of the populations that OMD finds arguably a better solution, given the allowed number of policy evaluations. The computed solution consists in covering the target distribution, up to the necessary connections between the squares and an inevitable density around the initial state.

\begin{figure}[tbh]
    \centering
    \begin{minipage}{.3\linewidth}
    \includegraphics[width=\linewidth]{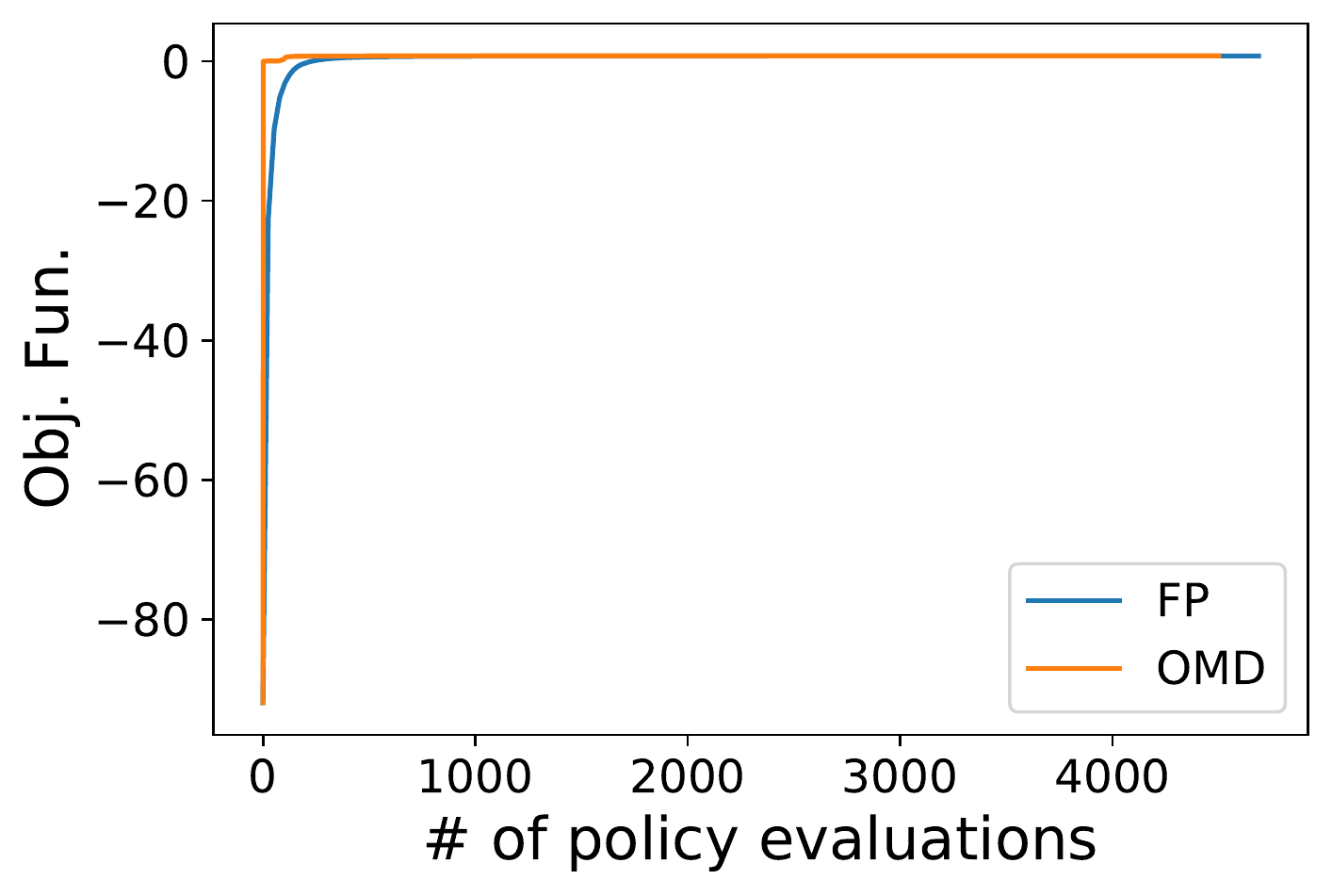}
    \end{minipage}
    \begin{minipage}{.3\linewidth}
    \includegraphics[width=\linewidth]{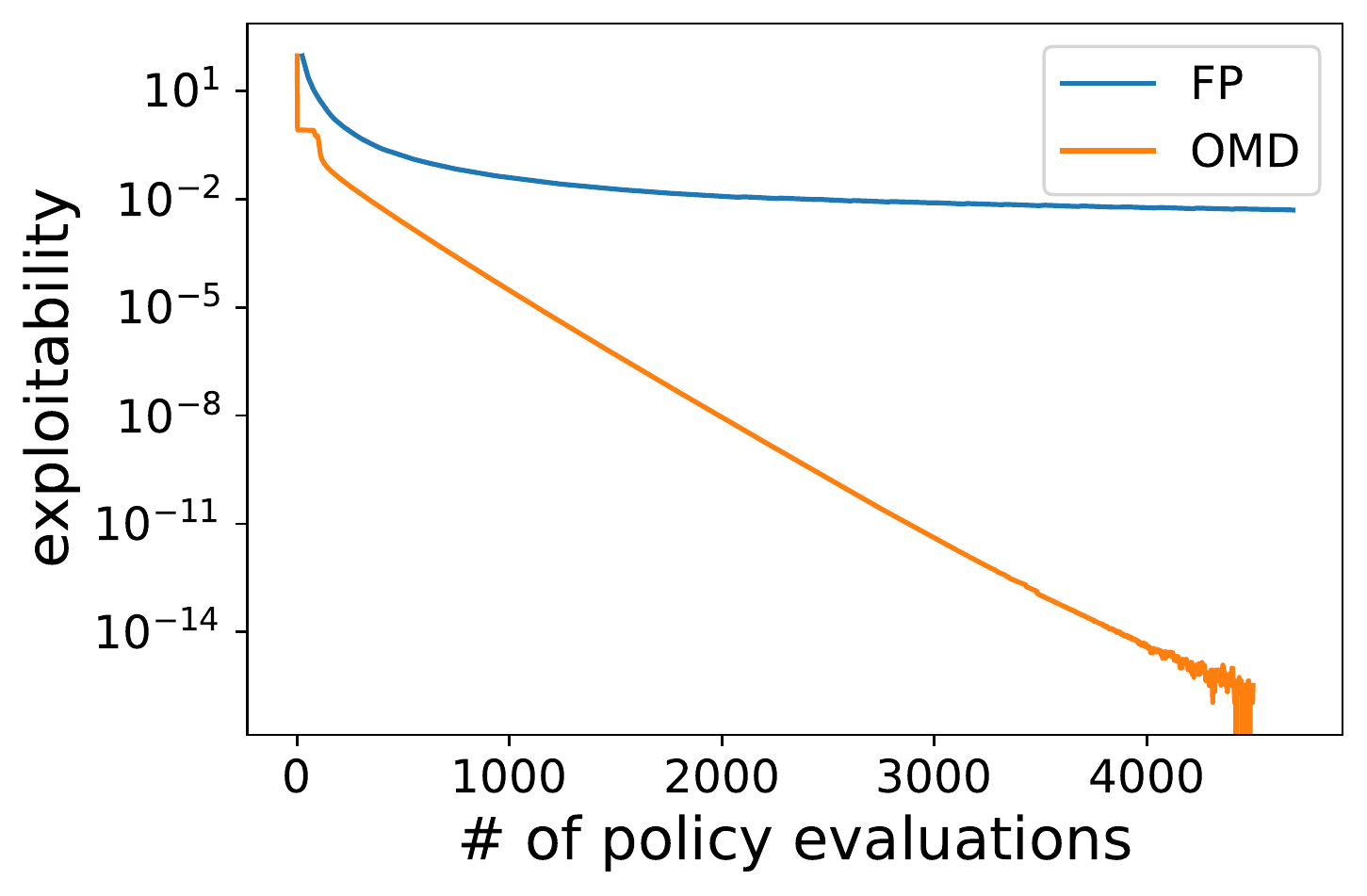}
    \end{minipage}
    \begin{minipage}{.19\linewidth}
    \includegraphics[width=\linewidth]{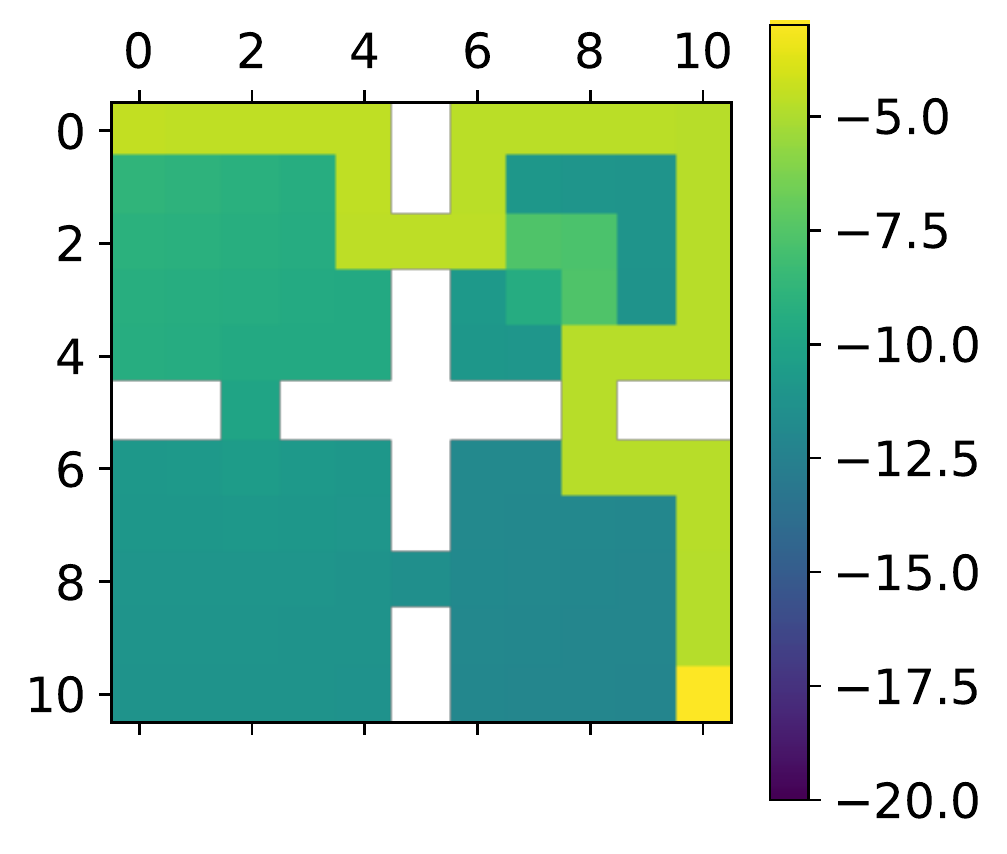}
    \end{minipage}
    \begin{minipage}{.19\linewidth}
    \includegraphics[width=\linewidth]{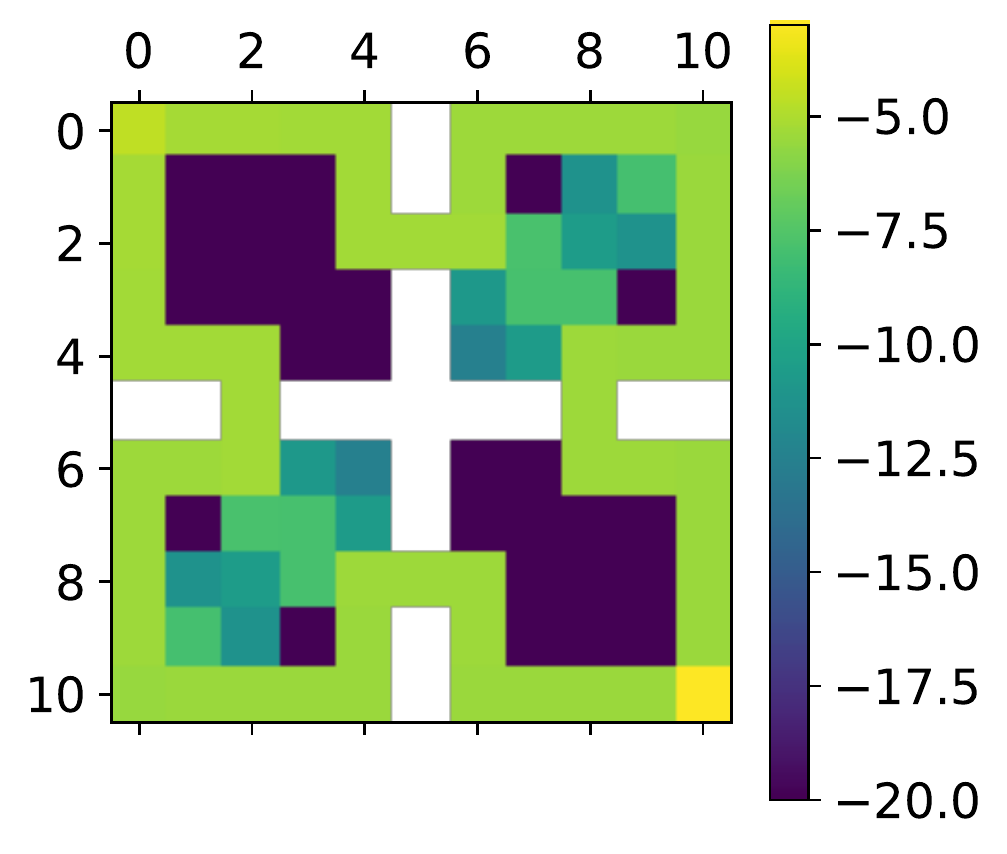}
    \end{minipage}
    \caption{\textbf{Constrained MDP.} 
    Reading order: \textbf{(a)} objective function; \textbf{(b)} exploitability;  \textbf{(c)}~log-density for the FP population; \textbf{(d)} log-density for the OMD population.
    }
    \label{fig:cmdp}
\end{figure}

\textbf{Constrained MDP.}
Here, the optimized function is $F_\text{cmdp}(\mu) = \langle \mu, r\rangle - \frac{\lambda}{2} \left(\max(0,\langle \mu, c\rangle)\right)^2$, with corresponding reward $\rc_\text{cmdp}(\cdot,\mu) = \nabla F_\text{cmdp}(\mu) = r - \lambda c \max(0,\langle \mu, c\rangle)$. 
Results are presented Fig.~\ref{fig:cmdp}, and one can make similar observations regarding FP and OMD (even though the difference of performance is much more noticeable through the exploitability here). There are multiple Nash in this case. FP favors one mode because best responses are computed using policy iteration (mixture policies are deterministic), while OMD solution is multimodal thanks to the policy being softmax.

\begin{figure}[tbh]
    \centering
    \begin{minipage}{.3\linewidth}
    \includegraphics[width=\linewidth]{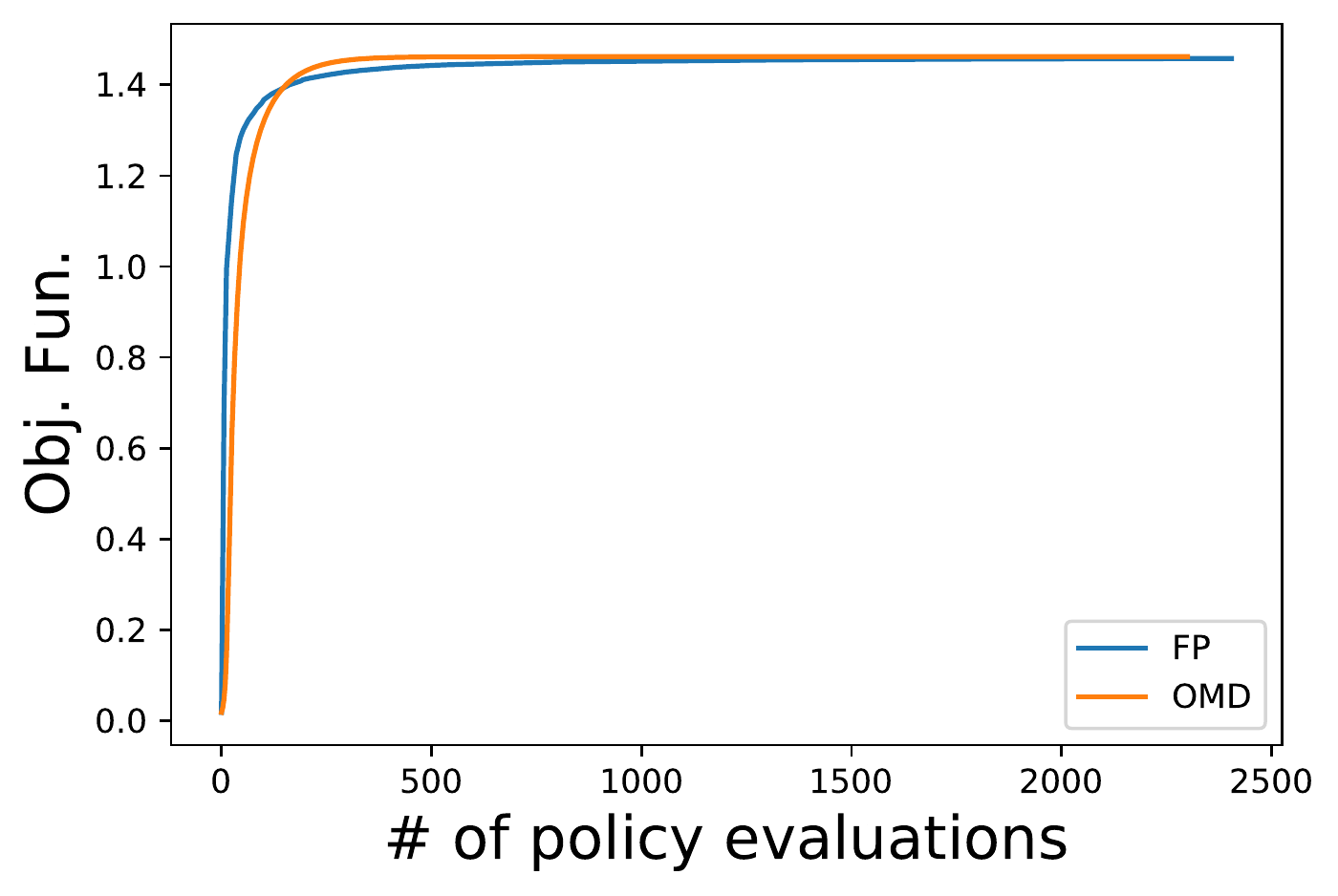}
    \end{minipage}
    \begin{minipage}{.3\linewidth}
    \includegraphics[width=\linewidth]{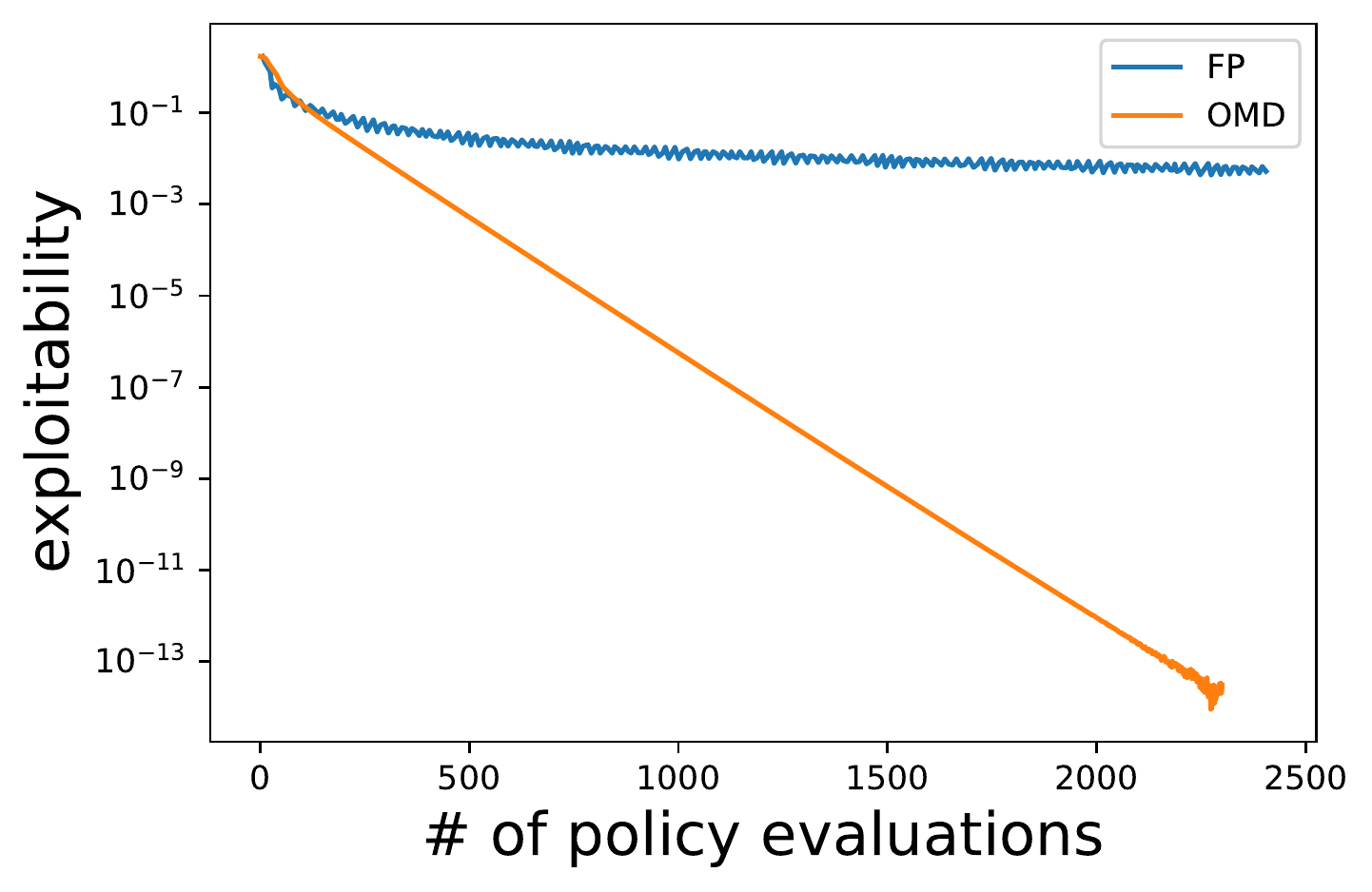}
    \end{minipage}
    \begin{minipage}{.19\linewidth}
    \includegraphics[width=\linewidth]{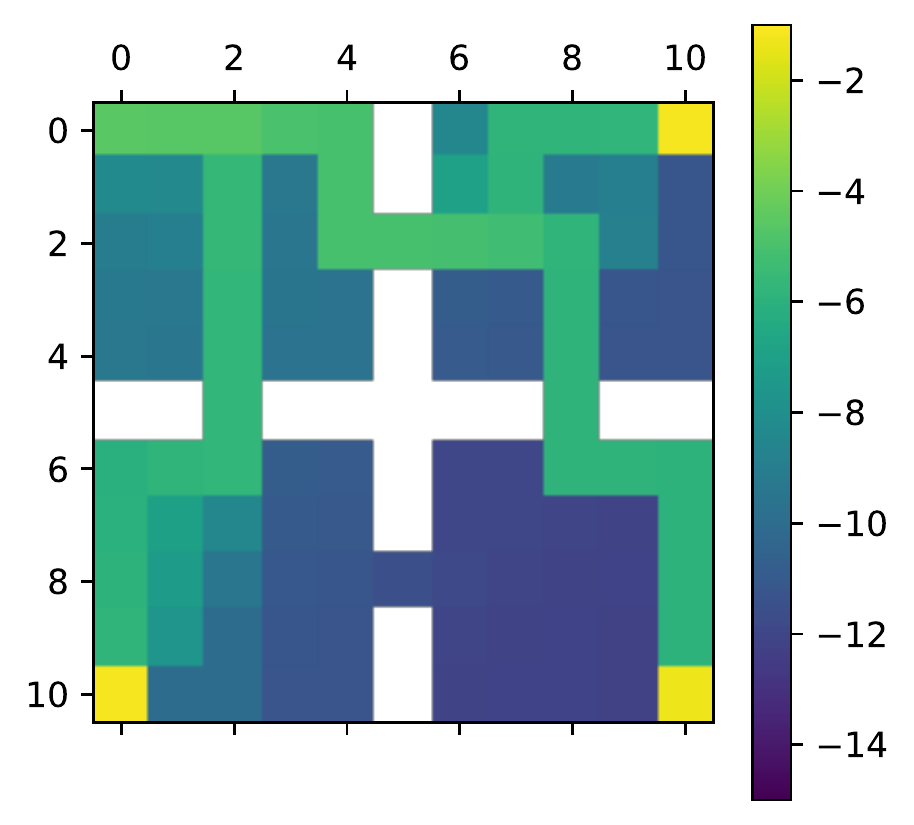}
    \end{minipage}
    \begin{minipage}{.19\linewidth}
    \includegraphics[width=\linewidth]{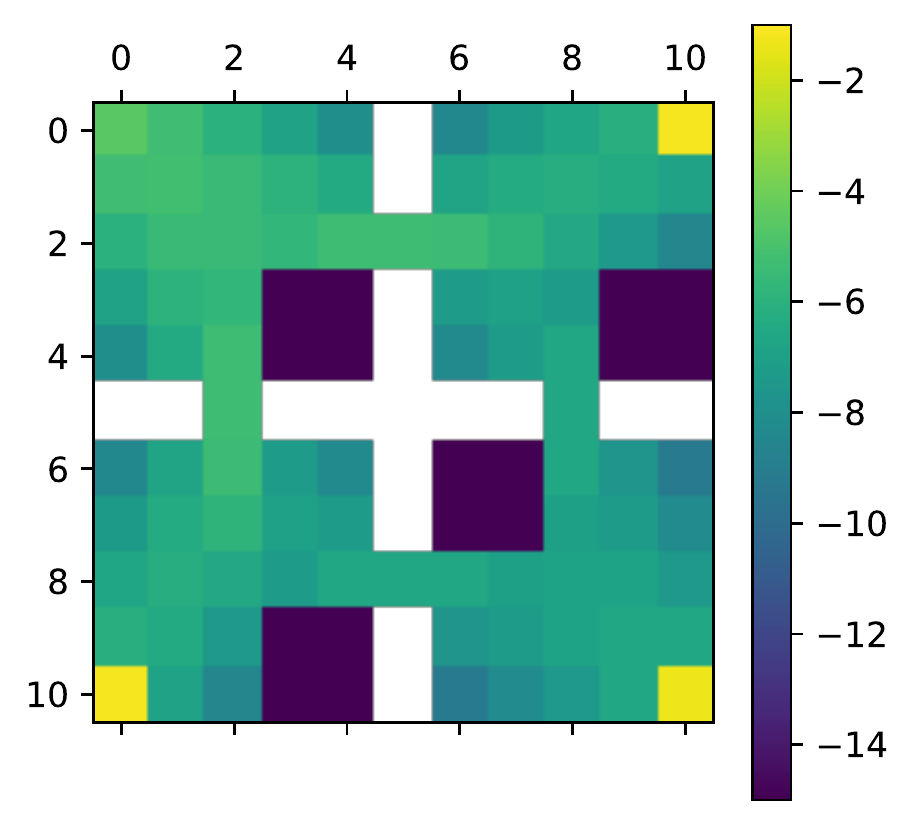}
    \end{minipage}
    \caption{\textbf{Multi-objective RL.} 
    Reading order: \textbf{(a)} objective function; \textbf{(b)} exploitability;  \textbf{(c)}~log-density for the FP population; \textbf{(d)} log-density for the OMD population.
    }
    \label{fig:morl}
\end{figure}

\textbf{Multi-objective RL.}
The function is $F_\text{morl}(\mu) = -\sum_{k=1}^3 (1-\langle \mu,r^k\rangle)^2$ and the reward $\rc_\text{morl}(\cdot,\mu) = \nabla F_\text{morl}(\mu) = 2\sum_{k=1}^3 r^k (1-\langle \mu, r^k\rangle)$. Results are presented Fig.~\ref{fig:morl}. Again, similar observations can be made. It might appears less clearly from the log-densities of the final population that OMD is closer to the Nash equilibrium, but it is the case both according to the objective function and to the exploitability (and very clearly here).

\textbf{Overall}, these numerical illustrations suggest that OMD, an approach specifically designed for MFGs, may be a viable alternative to \citet{hazan2019provably} approach for CURL problems. This highlights an additional evidence of the interest of bridging both communities.

\section{Conclusion}

We have shown that CURL is a subclass of MFGs and linked concepts from both fields: concavity and monotonicity, optimality conditions and exploitability, maximizer and Nash. We also discussed algorithms, showed that \citet{hazan2019provably} algorithm is indeed FP, and used this connection to FW to provide a rate of convergence for discrete-time FP. Our numerical illustrations suggest that it may be worth considering MFG algorithms for addressing CURL problems.

\looseness=-1
Solving an MFG from a global planner perspective leads to Mean Field Control (MFC) problems, which have also been considered for interpreting Deep learning algorithms~\citep{weinan2019mean}. MFC aims at maximizing $\langle \mu_\pi, \rc(\cdot,\mu_\pi)\rangle$ and, as such, may be seen as a special case of CURL, with possible additional difficulties (\textit{e.g.}, population-dependent dynamics or non-concave utility function). %
Thus, analytical and numerical solutions to MFC (\textit{e.g.},~\citep{pfeiffer2017numerical}) may also be useful for CURL. Another interesting perspective would be to study these algorithms in an approximate setting, both theoretically and empirically, something which is still quite preliminary in both fields, compared for example to RL.
Lastly, we have seen in Sec.~\ref{subsec:otherAlgs} that a part of the MFG literature (fixed-point-based) is not readily applicable to CURL. It would be interesting to study if these settings could be reconciled (adapting MFGs algorithms, or considering CURL in a stationary setting).

\newpage

\bibliographystyle{plainnat}
\bibliography{bib}

\end{document}